%% file: main.tex
\documentclass{article}

\usepackage[numbers]{natbib}
\usepackage[preprint]{neurips_2020}     
     
\usepackage{xcolor}
\usepackage[utf8]{inputenc} % allow utf-8 input
\usepackage[T1]{fontenc}    % use 8-bit T1 fonts
\usepackage[colorlinks=true, citecolor=black]{hyperref}       % hyperlinks
\usepackage{url}            % simple URL typesetting
\usepackage{booktabs}       % professional-quality tables
\usepackage{amsfonts}       % blackboard math symbols
\usepackage{nicefrac}       % compact symbols for 1/2, etc.
\usepackage{microtype}      % microtypography
\usepackage{amsmath}
\usepackage{graphicx}
\usepackage{textcomp}
\usepackage{gensymb}
\usepackage{bbm}
\usepackage{csquotes}
\usepackage{bm}
\usepackage{comment}
\usepackage{tabularx}
\usepackage{graphbox}
\usepackage{dcolumn}
\usepackage{textcomp}
\usepackage{hyperref}

\DeclareMathVersion{nxbold}
\SetSymbolFont{operators}{nxbold}{OT1}{cmr} {b}{n}
\SetSymbolFont{letters}  {nxbold}{OML}{cmm} {b}{it}
\SetSymbolFont{symbols}  {nxbold}{OMS}{cmsy}{b}{n}

\newcolumntype{d}[1]{D{.}{.}{#1}}

\newcolumntype{.}{D{.}{.}{-1}}
\makeatletter
\newcolumntype{B}[3]{>{\textfont0=\the{nxbold}\DC@{#1}{#2}{#3}}c<{\DC@end}}
\newcolumntype{J}[3]{>{\textfont0=\the@{#1}{#2}{#3}}c<{\DC@end}}
\makeatother
\newcommand\mc[1]{\multicolumn{1}{c}{#1}} % handy shortcut macro

\title{Probabilistic orientation estimation with matrix Fisher distributions}

\author{%
  David Mohlin\\
  KTH/Tobii\\
  \texttt{davmo@kth.se} \\
  \And
  Gerald Bianchi \\
  Tobii\\
  \texttt{Gerald.Bianchi@tobii.com} \\
  \And
  Josephine Sullivan \\
  KTH \\
  \texttt{sullivan@kth.se} \\
}

\begin{document}

\maketitle

\begin{abstract}
This paper focuses on estimating probability distributions over the set of 3D rotations ($SO(3)$) using deep neural networks. Learning to regress models to the set of rotations is inherently difficult
due to differences in topology between $\mathbb{R}^N$ and $SO(3)$. We overcome this issue by using a neural network to output the parameters for a matrix Fisher distribution since these parameters are homeomorphic to $\mathbb{R}^9$. By using a negative log likelihood loss for this distribution we get a loss which is convex with respect to the network outputs. By optimizing this loss we improve state-of-the-art on several challenging applicable datasets, namely Pascal3D+, ModelNet10-$SO(3)$ and UPNA head pose. All code used for this paper is available \href{https://github.com/Davmo049/Public_prob_orientation_estimation_with_matrix_fisher_distributions}{online}
\end{abstract}

\input{intro_related_work.tex}
\input{Method}

\input{ExperimentsAndResults}

\section{Conclusion \& Future work}
In this paper we have introduced a way to use neural networks to output probability distributions over the set of rotations with the matrix Fisher distribution. We show when optimizing the negative log likelihood of this distribution we end up with a convex loss. When applying this method on several datasets we get state-of-the-art performance. Our ablation studies show the relative robustness of the approach. 
% We have also introduced a type of preprocessing which allows for extracting image information for a certain region in an image which should be agnostic to the camera intrinsics and the bounding box shape.

Since the matrix Fisher distribution is unimodal it poorly models classes which have rotational symmetries. It could be interesting to try to create a loss supporting multimodal distributions while keeping the nice optimization properties of our loss.
It could be possible to use these estimated probabilities for time tracking filters such as the one described in \cite{lee2018bayesian}.

\section*{Broader Impact}

The methods described in this paper has obvious applications in fields which some consider ethically questionable such as for surveillance and military systems. One example could be determining heading for ships or airplanes for tactical planning. That being said, the orientation of objects is a fundamental property of objects in the real world and being able to accurately estimate this property should be helpful for many applications of either an ethically desirable or undesirable nature. In our opinion improving the techniques used for orientation estimation has a similar societal impact as improving the techniques used for classification or object detection.

The persons in the UPNA dataset are unlikely to be sampled from a uniform distribution of people across the world, for this reason one can not expect the reported performance to be accurate for the world population in general, that being said due to the small test size this reported performance might not reflect the average performance for any population. We do not believe this is an issue since models for predicting head pose which are deployed on a wider scale are very unlikely to use this dataset due to its small size and non-commercial licence. The method itself  is not reliant on any population specific feature.

\begin{ack}
%More information about this disclosure can be found at: %\url{https://neurips.cc/Conferences/2020/PaperInformation/FundingDisclosure}.
% TODO Josephine add additional information regarding your funding if needed.

The main author is an industrial PhD student at Tobii. This research is funded by scholarship from Wallenberg AI, Autonomous systems and Software Program (WASP). Computational resources was provided by KTH.

\end{ack}

\small
\bibliographystyle{plainnat}
\bibliography{refs}
\setcounter{section}{0}

\newpage

\vspace*{\fill}
\begingroup
\centering
{\huge Supplementary Material}
 
\endgroup
\vspace*{\fill}

\include{SupplementaryMaterial}

\end{document}

%% file: intro_related_work.tex
\section{Introduction}

Estimating the 3D rotation of an object from 2D images is one of the fundamental problems in computer vision. Several applications relying on 3D rotation estimation have been developed such as a robot grasping an object\cite{tremblay2018deep}, a self driving vehicle constantly sensing its surrounding environment \cite{Meng_2017}, an augmented reality system combining computer-generated information onto the real world \cite{marchand:hal-01246370}, or a system detecting the face orientation to enhance human-computer interactions \cite{Weidenbacher2006}.

Advances of deep learning techniques have resulted in improvements in estimation of 3D orientation. However, precise orientation estimation remains an open problem. The main problem is that the space of all 3D rotations lies on a nonlinear and closed manifold, referred to as the special orthogonal group $SO(3)$. This manifold has a different topology than unconstrained values in $\mathbb{R}^N$, where neural network outputs exist. As a result it is hard to design a loss function which is continuous without disconnected local minima. For example using euler angles as an intermediate step causes problems due to the so-called gimbal lock. Quaternions have a double embedding giving rise to the existence of two disconnected local minimas. Some more complicated methods use Gram-Schmidt \cite{zhou2019continuity} which has a continuous inverse, but the function is not continuous with a discontinuity when the input vectors do not span $\mathbb{R}^3$.

Despite these issues various deep learning based solutions have been suggested. One approach is to use one of the rotation representations and model the constraint in the loss function or in the network architecture \cite{mahendran20173d}. An alternative is to construct a mapping function, which directly converts the network output to a rotation matrix \cite{zhou2019continuity}.

Quantifying the 3D orientation uncertainty when dealing with noisy or otherwise difficult inputs is also an important task. Uncertainty estimation provides valuable information about the quality of the prediction during the process of decision making. Only recent efforts have been made on modeling the uncertainty of 3D rotation estimation \cite{prokudin2018deep, gilitschenski2019deep}. However, those methods still rely on complex solutions to fulfil the constraints required by their parameterization.

In this paper, we instead propose a deep learning approach to estimate the 3D rotation uncertainty by using the matrix Fisher probability density function developed in the field of directional statistics \cite{directional_statistics}. This unimodal distribution has been selected because of its relevant properties in regards to the problem of 3D orientation estimation: i) The parameterization is unconstrained so there is no need for complex functions to enforce constraints. ii) It is possible to create a loss for this distribution which has desirable properties such as convexity iii) The mode of the distribution can subsequently be estimated along with the uncertainty around that mode for further analysis.

Our method offers a simple solution to the problem of 3D orientation estimation, where a neural network learns to regress the parameters of the matrix Fisher distribution. While several other losses used for rotation estimation are discontinuous with multiple local minimas, with respect to the network output. In this paper we instead propose a loss which is convex with bounded gradient magnitudes, resulting in a stable training. In addition, we suggest a solution to efficiently compute the non-trivial normalizing constant of the distribution. Finally, the proposed method is evaluated on multiple problem domains and compared with the latest published approaches. The results show that our method outperforms all previous methods.

Our contributions include: 1) a method for estimating a probability distributions over the set of rotations with neural networks by using the matrix Fisher distribution, 2) a loss associated with this distribution and show it is convex with bounded gradients, and 3) an extensive analysis encompassing several datasets and recent orientation estimation works, where we demonstrate the superiority of our method over the state-of-the-art.

\section{Related Work}

3D rotation estimation has been studied over the last two decades. A common method estimates 3D rotation by aligning two sets of 3D feature points where each data set is matched and defined in a different coordinate system \cite{eggert1997estimating}. Another well-known approach matches 2D keypoints extracted from images with features of a known 3D model and recovers the 3D pose given the 2D-3D correspondences \cite{se2001vision, Lepetit09}
 
Recent methods using deep networks often predict 3D rotation directly from images without the knowledge of a 3D model of the object. Those methods can be grouped in two categories. The first one divides the set of rotations into regions and subsequently solves the 3D orientation estimation as a classification task. Subsequently, the classification network output is refined by a regression network \cite{mahendran2018mixed, liao2019spherical}.

The second category transforms the network output to a 3D rotation representation and learns to directly regress the 3D rotation given an image input. Commonly, quaternions \cite{xiang2017posecnn} or Euler angles \cite{mahendran20173d} representation are used.
However, the paper  \cite{zhou2019continuity} shows 
any rotation representation of dimensions four or lower is discontinuous, which  makes it difficult for the neural network to generalize over the set of rotations. 
They propose two continuous 5D and 6D rotation representations and construct a function that maps those representations to a rotation matrix. 
%Already cited above - Similarly, a specific activation function which converts the network output from real to rotation space has been presented in \cite{liao2019spherical}.

Recently some studies have investigated the prediction of rotation uncertainty using probability distributions over rotations. In \cite{prokudin2018deep} the parameters of a mixture of von Mises distribution using a biternion network are estimated. In \cite{gilitschenski2019deep}, the Bingham distribution over quaternions is used to jointly estimate a probability distribution over all rotation axes. However, their parameters have to be positive semidefinite due to their choice of probability distribution.

In this paper, we propose a solution which learns to regress the probability distribution with unconstrained parameters leading to a simple formulation of the problem of 3D rotation estimation.

%% file: Method.tex
\section{Method}
We train a neural network to estimate the 3D pose of an input image. %input signal. 
Specifically, the network outputs the parameters of the matrix Fisher distribution, which is a distribution over $SO(3)$.
From the predicted parameters we can obtain the maximum likelihood estimate of the input's 3D pose. In the rest of this section we review the matrix Fisher distribution and provide some visualizations to help the reader's interpretation of its parameters. Then we derive the loss, based on maximizing the likelihood of the labelled data, and finally explain how we deal with the distribution's complex normalizing constant when we calculate our loss and calculate the gradient during back-propagation. 

%colloquial - nasty normalizing constant when we calculate our loss and calculate the gradient during back-propagation training. 

%% In this paper we introduce the concept of a neural network outputting the parameters of a probability distribution over the set of 3D rotation matrices. In this section we describe the mathematical details of how we do this as well as the loss function we use for training. We begin by first 

\subsection{The matrix Fisher distribution on $SO(3)$}
%In this paper we introduce the concept of having a network output the parameters of a probability distribution over the set of rotation matrices. 
We model 3D rotation matrices probabilistically with the matrix Fisher distribution \citep{downs1972orientation,Khatri:rss:77}. This distribution has probability density function
\begin{equation}
    p(R \mid F) = \dfrac {1} {a(F)}\exp(\text{tr}(F^TR))
    \label{eqn:von_mises_pdf}
\end{equation}
where $F$ is an unconstrained matrix in $\mathbb{R}^{3\times3}$ parametrising the distribution, $R \in SO(3)$, and $a(F)$ is the distribution's normalizing constant. We will denote that $R$ is distributed according to a matrix Fisher distribution with $R \sim \mathcal{M}(F)$. The distribution is unimodal but visualizing the distribution in equation (\ref{eqn:von_mises_pdf}) is hard as it has a 3D domain. Fortunately, \citep{lee2018bayesian} describes a helpful visualization scheme, see figure \ref{fig:von_mises_vis} for details, which we use throughout the paper.

%nice visualization scheme, see figure \ref{fig:von_mises_vis} for details, which we use throughout the paper.

%% trim=left bottom right top,
\def\pwid{.18\linewidth}
\begin{figure}[t]
  \begin{tabular}{*{4}{m{\pwid}m{.01\linewidth}}}
    \includegraphics[width=\linewidth,trim=170 120 140 60,clip]{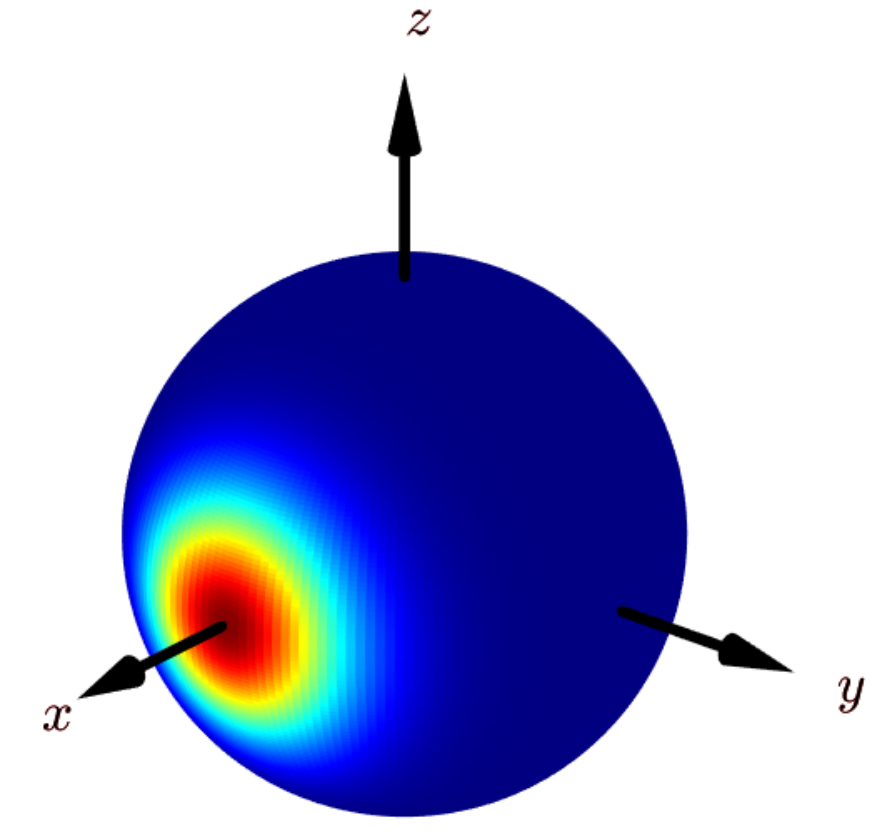} &
   {\vspace*{20pt}$+$} &
    \includegraphics[width=\linewidth,trim=170 120 140 60,clip]{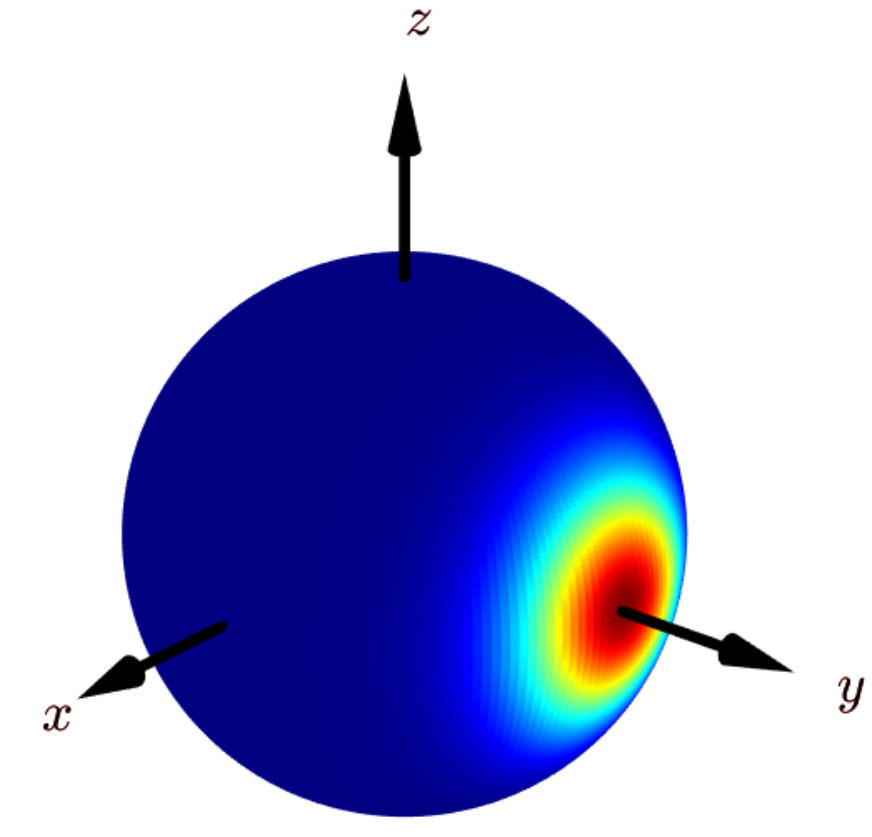} &
    {\vspace*{20pt}$+$} &
    \includegraphics[width=\linewidth,trim=170 120 140 60,clip]{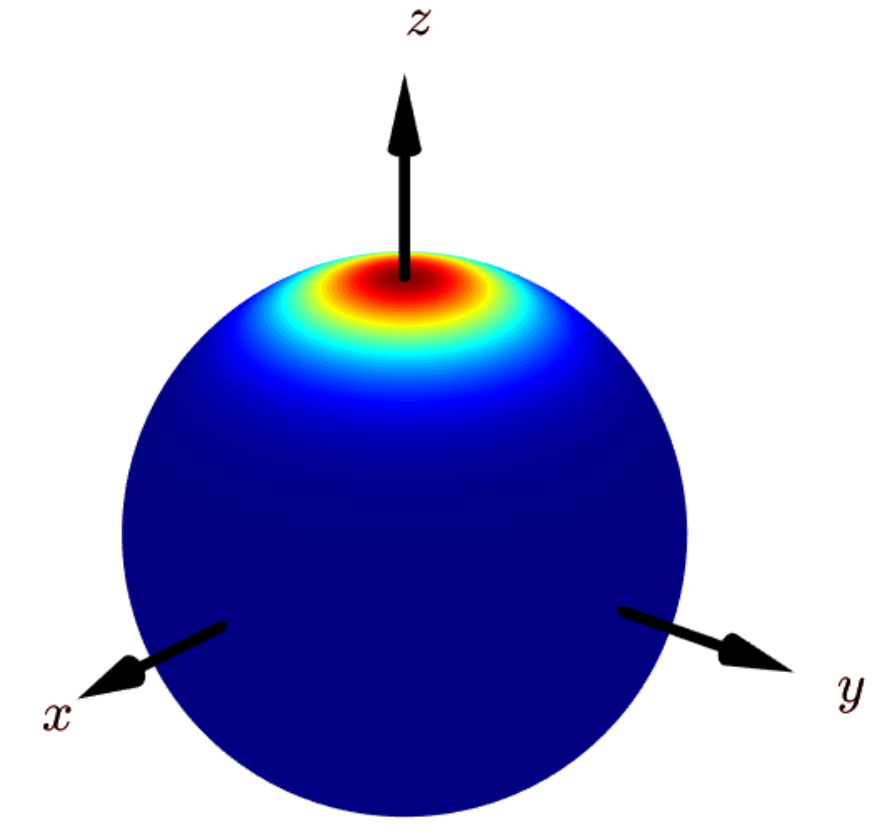} &
    {\vspace*{25pt}$=$} &
    \includegraphics[width=\linewidth,trim=170 120 140 60,clip]{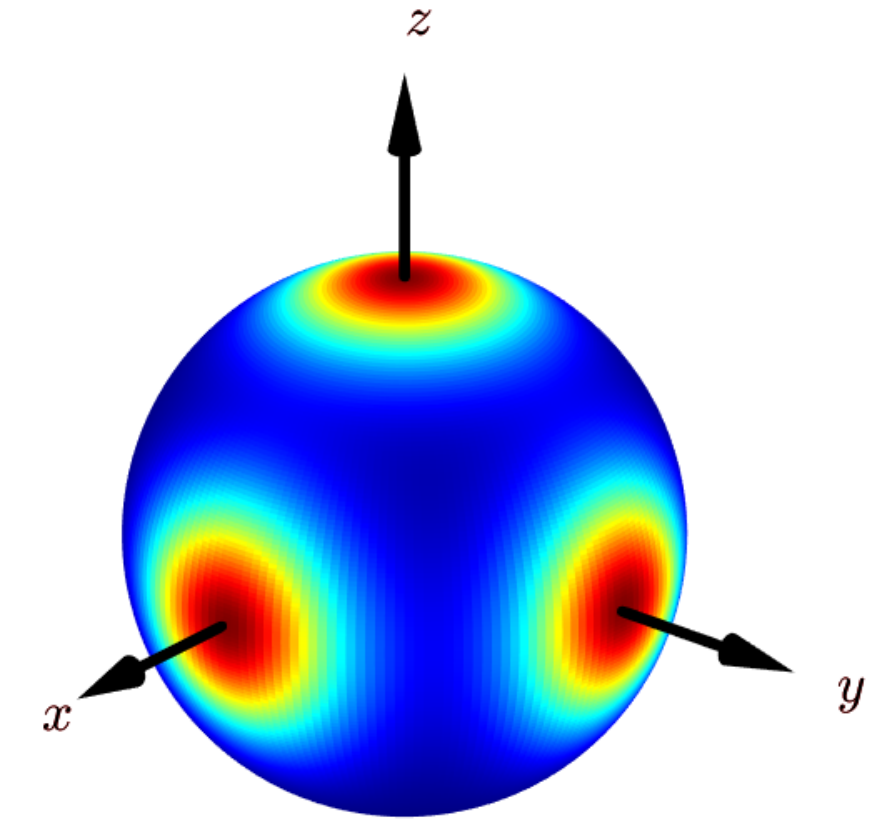}\\
                    \multicolumn{1}{c}{\footnotesize (a) $p(R e_1 \mid F)$} & &
                    \multicolumn{1}{c}{\footnotesize (b) $p(R e_2 \mid F)$} & &
                    \multicolumn{1}{c}{\footnotesize (c) $p(R e_3 \mid F)$} & &
                    \multicolumn{1}{c}{\footnotesize (d) Compact Vis.} 
                    
  \end{tabular}
  \caption[]{\textbf{Visualizing the matrix Fisher distribution on $SO(3)$.} We follow the convention of \citep{lee2018bayesian} and recreate their figures to explain the approach, similarly for figure \ref{fig:von_mises_spread}. For the above plots the parameter matrix is $F=\text{diag}(5, 5, 5)$. Let $e_1, e_2$ and $e_3$ correspond to the standard basis of $\mathbb{R}^3$ and is shown by the black axes. (a) This plot shows the probability distribution of $R e_1$ when $R \sim \mathcal{M}(F)$. Thus the pdf shown on the sphere corresponds to the probability of where the $x$-axis will be transformed to after applying $R \sim \mathcal{M}(F)$. (b) and (c) Same comment as (a) except consider $e_2$ and $e_3$ instead of $e_1$. (d) A compact visualization of the plots in (a), (b) and (c) is obtained by summing the three marginal distributions and displaying them on the 3D sphere. All four plots are plotted within the same scale and a \textit{jet} colormap is used.}
  \label{fig:von_mises_vis}
\end{figure}

\begin{figure}[htpb]
  \centering
  \def\phei{.12\textheight}
  \begin{tabular}{cccccc}
    \includegraphics[height=\phei,trim=170 120 140 60,clip]{images/VonMisesExamples/von_mises_1_all.pdf} &
    \includegraphics[height=\phei,trim=170 120 140 60,clip]{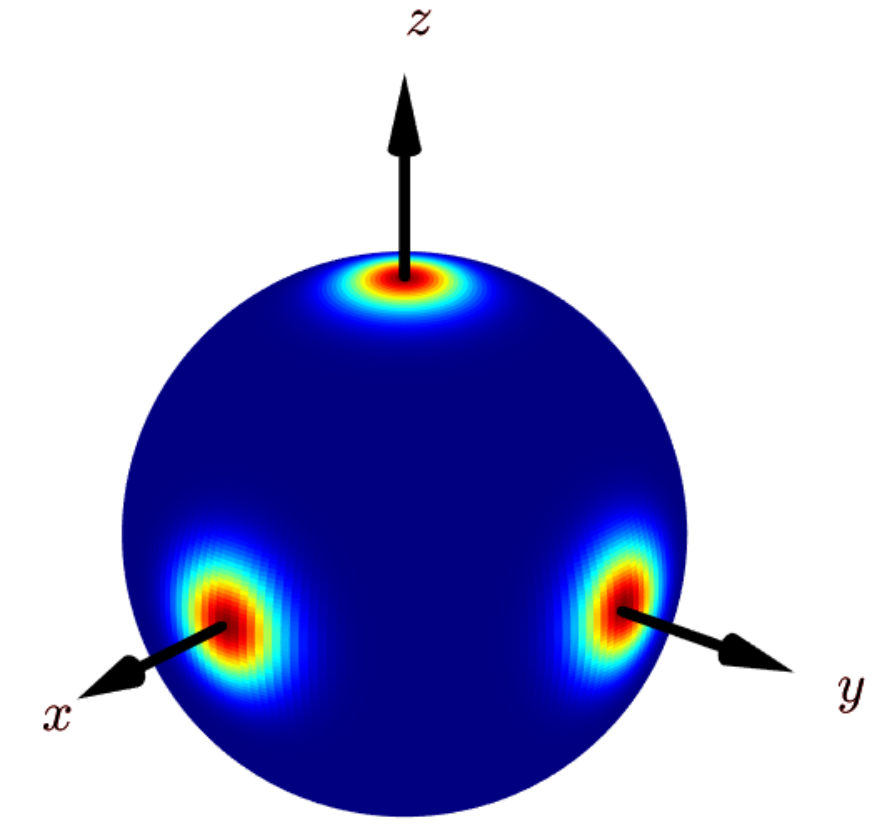} &
    \includegraphics[height=\phei,trim=170 120 140 60,clip]{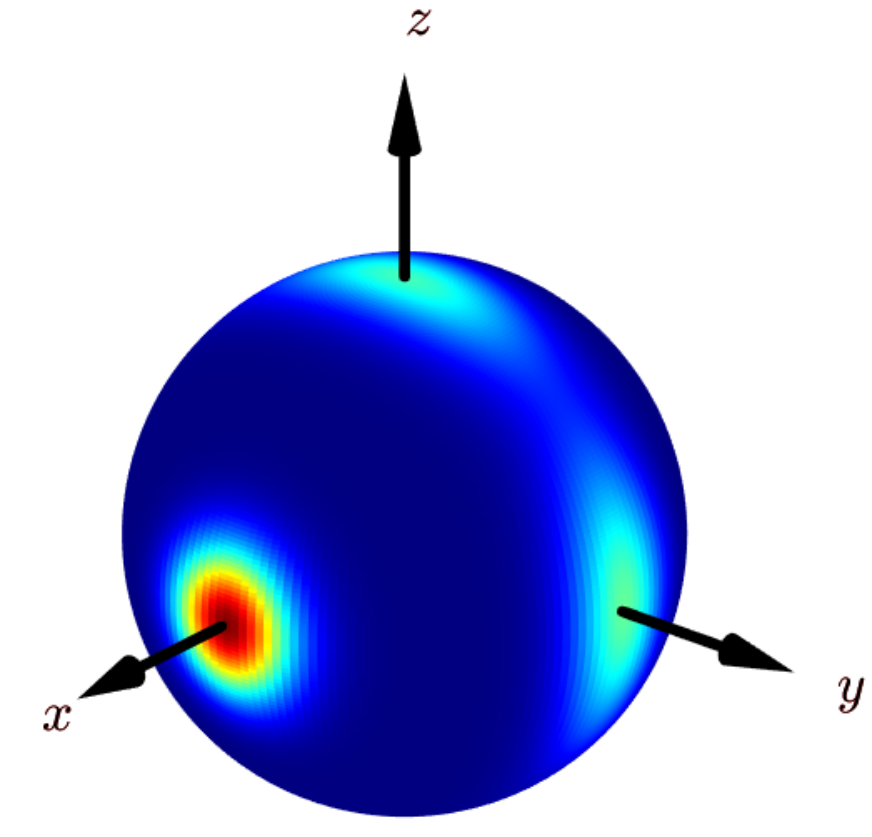} &
    \includegraphics[height=\phei,trim=130 120 140 60,clip]{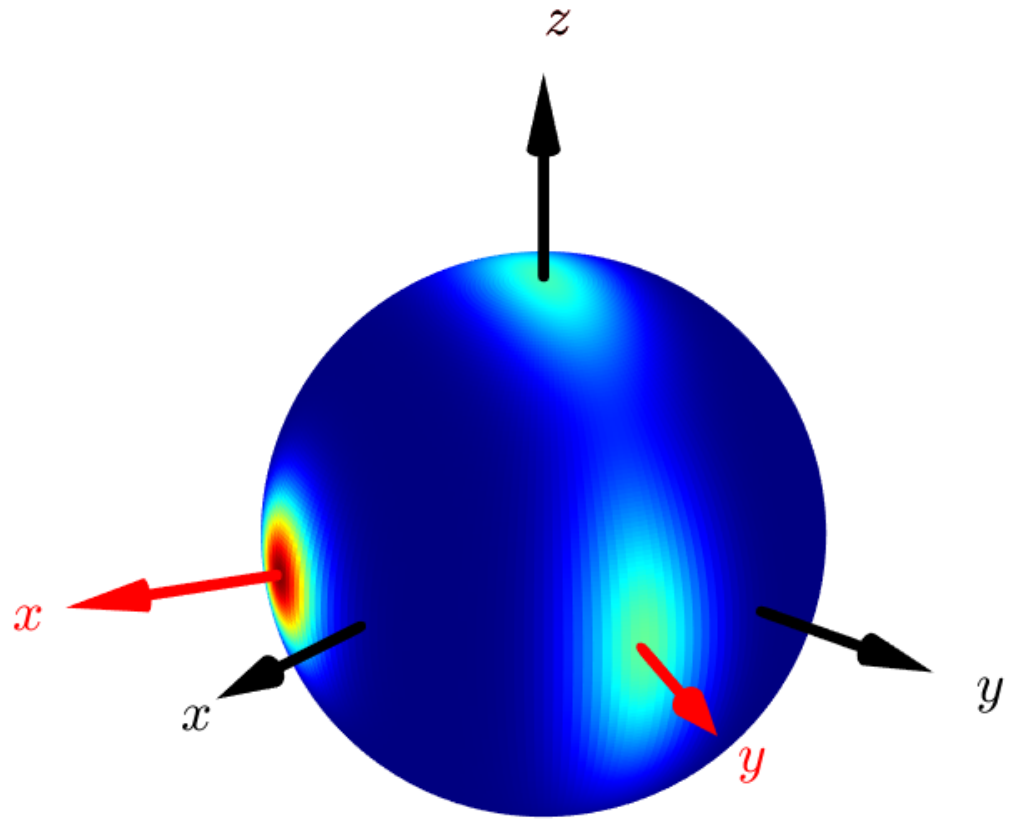} \\
                    {(a) $\text{diag}(5, 5, 5)$} &
                    { (b) $\text{diag}(20, 20, 20)$} &
                    { (c) $\text{diag}(25, 5, 1)$} &
                    { (d) $A\, \text{diag}(25, 5, 1)$}
  \end{tabular}
  \caption[]{ \textbf{Visualization of the matrix Fisher distribution for simple $F$ matrices}. (a) For a spherical $F$ the mode of the distribution is the identity. The distribution for each axis is circular and identical. (b) Here the axis distributions are more peaked than in (a) as the singular values are larger. (c) The distribution for the $y$- and $z$-axes are more elongated than for the $x$-axis as the first singular value dominates. (d) $A$ is the rotation matrix obtained by rotating around the $z$-axis by $-\pi/6$ degrees and thus the mode rotation is $A$ shown by the red axes. The shape of the axes distributions though remains as in (c).}
  \label{fig:von_mises_spread}
\end{figure}

Also not immediately apparent is how the shape of the distribution varies as $F$ varies. From \citep{eggert1997estimating} we know the mode of the distribution can be computed from the singular value decomposition of $F= U S V^T$, where the singular values are sorted in descending order, and setting
\begin{align}
    \hat{R} = U
    \begin{bmatrix}
    1 & 0 & 0 \\
    0 & 1 & 0 \\
    0 & 0 & \text{det}(U\,V)
    \end{bmatrix} V^T
    \label{eqn:von_mise_mode}
\end{align}
The latter operation ensures $\hat{R}$ is a proper rotation matrix. Similar results are available in \citep{downs1972orientation}. Figure \ref{fig:von_mises_spread} displays examples of the distribution for simple $F$ matrices. These figures show that larger singular values correspond to more peaked distributions.  To further help understanding of how the shape of the distribution relates to $F$ please consult section 3 of the supplementary material and \cite{lee2018bayesian}.

Finally, the deceptively simple form of the pdf in equation (\ref{eqn:von_mises_pdf}) hides the fact that its normalizing constant, $a(F)$, is rather complicated and non-trivial to compute accurately and efficiently. Explicitly the normalizing constant is 
\begin{equation}
  a(F) = \int_{R \in SO(3)} \exp(\text{tr}(F^TR))\, dR = \ {_1}F_1\left(\tfrac{1}{2}, 2, \Lambda_1\right) 
  \label{eqn:normalizing_factor}
\end{equation}
where ${}_1F_1(\cdot, \cdot, \cdot)$ is the generalized hypergeometric function of a matrix argument and
\begin{align}
  \Lambda_1 = \text{diag}(s_1-s_2-s'_3, s_2-s_1-s'_3, s'_3-s_1-s_2, s_1+s_2+s'_3)
\end{align}
Since this matrix is not positive semidefinite we use the extension from \cite{bingham1974antipodally} to define the function for all real matrices.
Here $s_1, s_2, s_3$ are the singular values of $F = USV^T$ and $s'_3 = s_3\text{det}(UV)$. Equation (\ref{eqn:normalizing_factor}) can be derived using results from \citep{jupp1979maximum} and a derivation appears in \citep{lee2018bayesian}.

\subsection{A negative log-likelihood loss function}

%% We train a neural network to take an input $x$ and output an estimate of the parameter matrix $F_x$. Thus it is necessary to define a loss function measuring the compatibility between $F_x$ and the ground truth rotation matrix $R_x$ for $x$. 
Assume we have a labelled training example $(x, R_x)$ where $x$ is the input and $R_x \in SO(3)$ its ground truth 3D rotation matrix. To train a neural network that estimates $F_x$ for input $x$, it is necessary to define a loss function measuring the compatibility between $F_x$ and $R_x$. As the pdf in equation (\ref{eqn:von_mises_pdf}) has support in all of $SO(3)$, we use  the negative log-likelihood of $R_x$ given $F_x$ as the loss:
\begin{align}
  \mathcal{L}(F_x, R_x) = -\log(p(R_x \mid F_x)) = \log(a(F_x)) - \text{tr}(F_x^T R_x)
  \label{eqn:loss}
\end{align}
%This loss has several nice properties such as it is
This loss has several interesting properties such as it is Lipschitz continuous, convex and has Lipschitz continuous gradients which makes it suitable for optimization. See supplementary material section 4 for proofs.

In practice the loss in equation \ref{eqn:loss} has an equilibrium far from the origin, in some experiments we believe this led to instability. To alleviate this problem we used a regularizing term which was 5\% larger than what is analytically correct to move the equilibrium closer to the origin.

%In practice the loss in equation (\ref{eqn:loss}) has an equilibrium between the estimated uncertainty and actual errors which is very far away from the origin, in some experiments we believe this led to instability. To alleviate this problem we used a regularizing term which was $1.05a(F)$ instead of $a(F)$ to move this equilibrium closer to the origin.
%where $R$ is the ground truth orientation.

%% We can differentiate this with respect to $F_x$ and get
%% \begin{gather}
%%     \nabla_{F_x} \text{Loss}(F_x, R) = -\nabla_{F_x} \log(a(F_x)) - R
%% \end{gather}
%One interpretation of the loss is the optimization tries to get the axes of rotation encoded in $F$ to point in the same directions as as those encoded by $R$.  but with a regularizing term which tries to keep the norm of $F$ small, which corresponds to not outputting too confident predictions.
%% This mapping and loss has several good properties which makes it suitable for optimization. For example the loss is Lipschitz continuous with connected sublevel sets.

\subsection{Efficiently estimating the normalizing constant}

%The normalizing constant $a(F)$ is tricky to evaluate
The normalizing constant $a(F)$ is non-trivial to evaluate
accurately and quickly. To allow computationally feasible training we fit a simple function to approximate ${_1}F_1\left({1}/{2}, 2, \cdot \right)$ where the input has the form of $\Lambda_1$ and also to its derivative w.r.t to the inputs $s_1, s_2, s'_3$. The latter approximating functions are used for back-prop training, while the former approximation is used to visualize the loss during training. We used software from \citep{koev2006efficient} to compute the value ${_1}F_1\left({1}/{2}, 2, \Lambda_1\right)$ and its numerical gradients for multiple values of $\Lambda_1$. Via a process of manual inspection and curve fitting of simple functions we create an approximation of this function. The exact details are given in section 5 of the supplementary material. 

Though perhaps our approach is far from optimal it does make training and testing computationally feasible and our results indicate that our approximation is sufficiently accurate to allow us to train a powerful network. Some preliminary experiments would also seem to indicate that the results are not so sensitive to the accuracy of the approximation, though having a more theoretically well-founded approach would be more satisfying such as adapting the approach in \citep{gilitschenski2019deep} or approximating the relevant analytical derivatives calculated in \citep{lee2018bayesian}. 

%% In the appendix we show for the function
%% ${_1} F_1: \mathbb{R} \times \mathbb{R} \times \mathbb{R}^4 \rightarrow \mathbb{R}$ has the following property.
%% \begin{equation}
%%     \sum\limits_{i=1}^{4} \nabla_{\Lambda_i} {_1} F_1\left(\tfrac{1}{2}, 2, \Lambda\right) = 1
%% \end{equation}
%% Here we use a slightly different notation than what is conventional, the common notation is to have a full matrix as the third argument, but the output is only dependent on the singular values. Therefore we instead use a vector of the singular values as the argument.

%% We then use the software from \cite{koev2006efficient} to compute the value of this function and its numerical gradients for a few points accurately.
%% We then fit a simplified function to these values to get

%% \begin{equation}
%%     \nabla_{\Lambda} log({_1}F_1(\frac{1}{2}, 2, \Lambda))
%% \approx
%%     \frac
%%       {1 + \Lambda}
%%       {\sum\limits_{i=1}^{4} 1 + \Lambda_i}
%% \end{equation}

%% The remaining factors of $a(F)$ have trivial gradients which means backpropagation can be performed using these results and approximations.

%% file: ExperimentsAndResults.tex
\section{Experimental Details}

We test our proposed approach on three separate datasets Pascal3D+,
ModelNet10-$SO(3)$ and UPNA head pose. We briefly describe these
datasets, the pre-processing we applied to the images from each
dataset before training and then the evaluation metrics.

\subsection{Datasets \& Pre-processing}
%% \subsubsection*{Pascal3D+}

\textbf{Pascal3D+} \citep{xiang_wacv14} has 12 rigid object classes and contains images from Pascal VOC and ImageNet of these classes. Each image is annotated with the object's class, bounding box and 3D pose. The latter is found by having an annotator align a 3D CAD model to the object in the image.
%\textbf{Pascal3D+} \citep{xiang_wacv14} is a dataset containing instances from 12 classes appearing in images from ImageNet and Pascal VOC. Each image is annotated with the object's class, bounding box and the orientation and 3d position. The latter is found by having an annotator align a 3d CAD model to the object in the image.

We pre-process each image by applying a homography so that the transformed image appears to come from a camera with known intrinsics and a principal axis pointing towards the object. This approach is similar to \citep{linden2018appearance} and \citep{zhang2018revisiting}. More details are given in section 7 of the supplementary material. We perform data augmentation similar to the data augmentations introduced in \citep{mahendran20173d}, but adapted for our preprocessing. At test time we apply the same type of homography transformation as applied during training, but no data augmentation.
%but no test time augmentation is performed.

%% For data augmentation we perturb the bounding box coordinates. The effect of this is similar to applying a scaling, translation and a small homography transformation. We also perform a horizontal flip with 50\% probability and small random rotations similar to the data augmentations introduced in \citep{mahendran20173d} but for a different kind of pre-processing.  For all augmentations we keep the up direction the same for the augmented and the original input image. 

%% \subsubsection*{ModelNet10-SO(3)}

\textbf{ModelNet10-$\bm{SO(3)}$}\citep{liao2019spherical} is a synthetic dataset. It is created by rendering rotated 3D models from ModelNet10 \citep{wu20153d} with uniformly sampled rotations.
%It is created by uniformly sampling rotations and then rendering 3d models from ModelNet10 \citep{wu20153d} with these rotations. 
The task is to estimate the applied rotation matrix.

We do not use any preprocessing for these images since the object is
already centered and of a reasonable size.  We do not use any data
augmentation as the original paper did not use any and we want a fair
comparison between the losses.

%% \begin{figure}[htpb]
%%   \centering
%%   \includegraphics[scale=0.5]{images/UPNA_example.pdf}
%%   \caption{Sample after preprocessing overlaid with ground truth orientation (blue forward, green up, red left)}
%% \end{figure}

\textbf{UPNA head pose} \citep{ariz2016novel} consists of videos with synchronized annotations of keypoints for the face in the image as well as its 3D rotation and position. The dataset has 10 people each with 11 recordings.

 From the keypoint annotations we create a face bounding box for each image. After this we perform a small random perturbation of this bounding box to degrade the quality of the bounding box to be similar to what one would expect to get from a face detector. Using this artificial bounding box enables us to use the same data augmentation and preprocessing as we used for Pascal3D+.
%% We first create bounding boxes centered at $\frac{\max\limits_i {p_x}_i + \min\limits_i {p_x}_i} {2}$, the center y coordinate is computed analogously. We choose the width to be proportional to $\max\limits_i {p_x}_i - \min\limits_i {p_x}_i$, analogously for the height. After this we perform a small random perturbation of this bounding box to degrade the quality of the bounding box to be similar to what one would expect to get from a face detector. Using this artificial bounding box enables us to use the same data augmentation and preprocessing as we used for Pascal3D+.
There is no official training/test split for this dataset. We create a split by using the last 2 people as the test set. We did not use a validation set since we did not apply a new hyperparameter search for the dataset. Our test split differs from \citep{gilitschenski2019deep} as theirs is not publicly available.

\subsection{Details of network \& training}

%% \begin{figure}[htpb]
%%   \centering
%%   \includegraphics[scale=0.75]{images/network_arch.png}
%%   \caption{Network architecture}
%% \end{figure}

We run experiments with ResNet-101 as our backbone network. The
ResNet-101 parameters are initialized from pre-trained ImageNet
weights. The object's class is encoded by an embedding layer that
produces a 32-dimensional vector and which is appended to the ResNet's
activations obtained from the final average pooling layer.  We apply 3
fully connected layers to this vector with $[512, 512, 9]$ nodes output at each layer.
%and the final network outputs 9 numbers.

We fine-tune the embedding and fully connected layer weights for 2 epochs. We use SGD and start with a learning rate of 0.01. We use a batch size of 32 and train for 120 epochs. For Pascal3D+ we reduce this learning rate by a factor 10 at epochs 30, 60 and 90. For ModelNet10-SO(3) we train for 50 epochs and reduce the learning rate by a factor of 10 at epochs 30, 40 and 45. For UPNA head pose we use the same hyperparameters as for Pascal3D+, except we do not use a class embedding since there are only faces in this dataset.

\subsection{Evaluation metrics}

The evaluation metrics used are based on the geodesic distance:
\begin{gather}
    d(R, \hat{R}) = \arccos\left(\frac{1}{2}\left(\text{tr}(R^T\hat{R})-1\right)\right)
    \label{eqn:eval_metric}
\end{gather}
where $R$ and $\hat{R}$ are the ground truth and estimated rotation respectively. This metric returns an angle error and we measure it in degrees. For a test set $\mathcal{X}$, containing tuples $(x, R_x)$ of input $x$ and its ground truth rotation $R_x$, we summarize performance on $\mathcal{X}$ with the median angle error and $\text{Acc}@Y$:
\begin{align}
\text{Acc}@Y = \frac{1}{|\mathcal{X}|} {\sum_{(x, R_x) \in \mathcal{X}} \mathbbm{1}(d(R_x, \hat{R}_x) < Y)}
\end{align}
where $\mathbbm{1}(\cdot)$ is the indicator function, $\hat{R}_x$ is the estimated rotation for input $x$ and $|.|$ is the cardinality of a set. To compute the overall performance on a dataset the median angle error and $\text{Acc}@Y$ are first computed per class and then averaged across all classes. For the UPNA dataset we use the mean geodesic error angle instead of the median to allow more direct comparison with the results in \citep{gilitschenski2019deep}. 

We have done all development and hyper-parameter optimization where the full training set was partitioned into a training and validation set. After hyper-parameter optimization, we have used the full training set for training and these are the numbers presented in the tables. For the Pascal3D+ dataset, we use the ImageNet validation split for the test set. Some samples of Pascal3D+ are labeled as \enquote{truncated}, \enquote{difficult} or \enquote{occluded}. We exclude these samples from our evaluations similar to other reported results \citep{liao2019spherical}. This implementation detail had only a very slight effect on performance. 

\section{Results}
\paragraph{Quantitative results}
Table \ref{tab:pascal_performance} compares the performance of our method and recent high performing approaches. 
Table \ref{tab:pascal_per_class_performance} compares per class performance for some classes on on Pascal3D+ with the previous state-of-the-art method \citep{mahendran2018mixed}.
Our method significantly outperforms all the prior approaches. When the training set is augmented with the synthetic dataset from \citep{su2015render}, we further reduce the mean over medians angle error by approximately 1 degree.

The results reported in table \ref{tab:modelnet_performance} show that our method also achieves state-of-the-art performance on ModelNet10-SO(3).
On the UPNA head pose dataset our algorithm gives a mean angle error of 4.5 degrees. This compares well to \citep{gilitschenski2019deep} who quote a performance of 6.3 degrees. Unfortunately, since different test splits were used a direct comparison is hard. We use a test split where there is no overlap between individuals in the test and training set as compared to \citep{gilitschenski2019deep} where the same individuals can appear in both the training and test sets (personal communication). Thus it is likely our test split is more challenging. Despite this we get a significantly lower average error.
\begin{table}[t]
  \centering
  \caption{\textbf{Performance on Pascal3D+}. Results are reported for the median angle error, Acc@$\pi/6$ and Acc@$\pi/12$. The last column indicates if the training set was augmented with the synthetic dataset from \cite{su2015render}.}
  \label{tab:pascal_performance}
  {
  \begin{tabular}{l*{3}{J{.}{.}{2.1}}c}
    \toprule    
\textbf{Method} & \mc{MedErr} & \mc{Acc@${\pi}/{6}$ (\%)} & \mc{Acc@${\pi}/{12}$ (\%)} & Use synth. \\
\midrule
\citet{mahendran20173d} & 15.38 & - & - & $\times$ \\
\citet{pavlakos20176} & 14.16 & - & - & $\times$ \\
\citet{tulsiani2015viewpoints} & 13.60 & 81.0 & - & $\times$ \\
\citet{su2015render} & 11.70 & 82.0 & - & \checkmark  \\
\citet{grabner20183d} & 10.90 & 83.9 & - & $\times$ \\
\citet{prokudin2018deep} & 10.40 & 83.9 & - & $\times$ \\
\citet{mahendran2018mixed} & 10.10 & 85.9 & - & \checkmark  \\
\citet{liao2019spherical} & \ 9.20 & 88.7 & - & \checkmark \\
\midrule
Ours & \ 8.90 & 90.8 & 74.5 & $\times$ \\
Ours &  \multicolumn{1}{B{.}{.}{2,1}}{8.10} & \multicolumn{1}{B{.}{.}{2,1}}{93.1} & \multicolumn{1}{B{.}{.}{2,1}}{78.2} & \checkmark \\
\bottomrule
  \end{tabular}
  }
\end{table}

\begin{table}[b]
\caption{\textbf{Pascal3D+ per-class performance} of our method, with or without using extra synthetic training data, compared to the competitive method \citet{mahendran2018mixed}. The top three rows report the median angle error per class measured in degrees. The bottom three rows report Acc@${\pi}/{6}$ measured as a percentage.}
\label{tab:pascal_per_class_performance}  
\centering
\begin{tabular}{l*{13}{J{.}{.}{2.1}}}
\toprule
\textbf{Method} & \mc{aero} & \mc{bike} & \mc{boat} & \mc{bottle} &  \mc{bus} & 
%\mc{car} & 
\mc{chair} & \mc{dtable} & 
%\mc{mbike} & 
\mc{sofa} & \mc{train} & 
%\mc{tv} & 
\mc{\textbf{mean}} \\
\midrule
\citep{mahendran2018mixed} & 8.5 & 14.8 &
20.5 & \multicolumn{1}{B{.}{.}{2,1}}{7.0} & \multicolumn{1}{B{.}{.}{2,1}}{3.1} &
%5.1 & 
9.3 & 11.3 & 
%14.2 &
10.2 & 5.6 & 
%11.7 & 
10.1 \\
Ours w/o & 10.2 & 14.7 & 12.6 & 8.2 & 3.5 & %\multicolumn{1}{B{.}{.}{2,1}}{3.8} &
7.8 & \multicolumn{1}{B{.}{.}{2,1}}{8.4} & 
%13.1 & 
7.7 & 5.4 & 
%11.4 & 
8.9\\
Ours with & \multicolumn{1}{B{.}{.}{2,1}}{6.8} & \multicolumn{1}{B{.}{.}{2,1}}{12.3} & \multicolumn{1}{B{.}{.}{2,1}}{12.2} & 7.6 & 3.8 & 
%3.9 & 
\multicolumn{1}{B{.}{.}{2,1}}{6.4} & 10.6 & %\multicolumn{1}{B{.}{.}{2,1}}{10.8} &
\multicolumn{1}{B{.}{.}{2,1}}{7.4} & \multicolumn{1}{B{.}{.}{2,1}}{5.3} & %\multicolumn{1}{B{.}{.}{2,1}}{10.3}&
\multicolumn{1}{B{.}{.}{2,1}}{8.1}\\
\midrule
\citep{mahendran2018mixed}  & 87.0 & 81.0 & 64.0 & \multicolumn{1}{B{.}{.}{2,1}}{96.0} & 97.0 & 
%95.0 & 
92.0 & 67.0 & %85.0 & 
97.0 & 82.0 & 
%88.0 & 
85.9 \\
Ours w/o & 87.9 & 82.3 & 77.8 & 95.5 & 98.4 & 
%98.3 &
94.8 & \multicolumn{1}{B{.}{.}{2,1}}{82.0} & 
%88.0 & 
98.1 & 97.9 & 
%89.0 & 
90.8 \\
Ours with & \multicolumn{1}{B{.}{.}{2,1}}{93.9} & \multicolumn{1}{B{.}{.}{2,1}}{89.2} & \multicolumn{1}{B{.}{.}{2,1}}{80.0} & 94.6 & \multicolumn{1}{B{.}{.}{2,1}}{98.6} & %\multicolumn{1}{B{.}{.}{2,1}}{99.1} &
\multicolumn{1}{B{.}{.}{2,1}}{98.7} & 
78.0 & 
%\multicolumn{1}{B{.}{.}{2,1}}{93.1} &
\multicolumn{1}{B{.}{.}{2,1}}{99.7} & \multicolumn{1}{B{.}{.}{2,1}}{98.4} & %\multicolumn{1}{B{.}{.}{2,1}}{94.2} &
\multicolumn{1}{B{.}{.}{2,1}}{93.1} \\
  \bottomrule
\end{tabular}  
\end{table}
%%%% ModelNet-SO(3) %%%%%
%The results reported in table \ref{tab:modelnet_performance} show that our method also achieves state-of-the-art performance on ModelNet10-SO(3). 
% The results are good
%than\citep{gilitschenski2019deep}.

\begin{table}[t]
  \caption{\textbf{Performance on ModelNet10-SO(3)}. * indicates the numbers reported in the original paper, but $\dagger$ denotes the revised numbers \citep{spherical_regression_code} where the evaluation metric uses the distance defined in equation (\ref{eqn:eval_metric}). Thus we compare the performance of our method to the latter numbers.}   
    \label{tab:modelnet_performance}
  \begin{center}
  \begin{tabular}{l*{5}{J{.}{.}{2.1}}}
    \toprule
Method & \mc{MedErr (deg)} & \mc{\textbf{Acc}@${\pi}/{6}$ (\%)} & \mc{Acc@${\pi}/{12}$ (\%)} & \mc{Acc@${\pi}/{24}$ (\%)} \\
\midrule
\citet{liao2019spherical}* &  20.3 & 70.9 & 58.9 & 38.4 \\
\citet{spherical_regression_code}$\dagger$ &  28.7 & 65.8 & 49.6 & 35.2 \\
Ours & \multicolumn{1}{B{.}{.}{2,1}}{17.1} & \multicolumn{1}{B{.}{.}{2,1}}{75.7} & \multicolumn{1}{B{.}{.}{2,1}}{69.3} & \multicolumn{1}{B{.}{.}{2,1}}{55.2} \\
 \bottomrule
  \end{tabular}
  \end{center}
\end{table}

\begin{table}[htpb]
\caption{\textbf{Per class performance on ModelNet10-SO(3)} of our method.}
\label{tab:modelnet_performance_per_class}  
  \renewcommand{\arraystretch}{1.1}
  % TODO fill in all values
  %% \begin{center}
  {\centering
  \begin{tabular}{l*{11}{J{.}{.}{2.1}}}
    \toprule
    \textbf{Metric} & \mc{bathtub} & \mc{bed} & \mc{chair} & \mc{desk} & \mc{dress.} & \mc{t.v.} & \mc{n. stand} & \mc{sofa} &  \mc{table} & \mc{toilet}\\
    \midrule
        MedErr & 89.1 & 4.4 & 5.2 & 13.0 & 6.3 & 5.8 & 13.5 & 4.0 & 25.8 & 4.0\\
    Acc@${\pi}/{6}$ & 40.3 & 90.8 & 93.5 & 67.4 & 73.9 & 86.3 & 61.4 & 94.4 & 51.1 & 98.1\\
    Acc@${\pi}/{12}$ & 32.2 & 88.2 & 88.1 & 53.6 & 68.2 & 79.0 & 51.6 & 91.9 & 44.6 & 95.7\\
%    MedErr$\downarrow$ & 89.1$\textdegree$ & 4.4$\textdegree$ & 5.2$\textdegree$ & 13.0$\textdegree$ & 6.3$\textdegree$ & 5.8$\textdegree$ & 13.5$\textdegree$ & 4.0$\textdegree$ & 25.8$\textdegree$ & 4.0$\textdegree$\\
 %   Acc@${\pi}/{6}  \uparrow$ & 40.3\% & 90.8\% & 93.5\% & 67.4\% & 73.9\% & 86.3\% & 61.4\% & 94.4\% & 51.1\% & 98.1\%\\
  %  Acc@${\pi}/{12} \uparrow$ & 32.2\% & 88.2\% & 88.1\% & 53.6\% & 68.2\% & 79.0\% & 51.6\% & 91.9\% & 44.6\% & 95.7\%\\
    \bottomrule
  \end{tabular}
  }
  %% \end{center}
\end{table}

\begin{figure}[!b]
\def\pwid{.16\linewidth}
\centering
  \begin{tabular}{*{4}{p{.16\linewidth}}}
    \multicolumn{1}{c}{\includegraphics[width=\pwid]{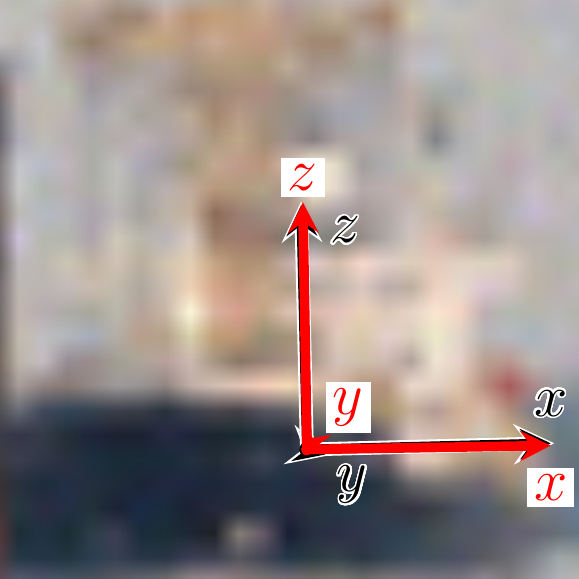}} &
    \multicolumn{1}{c}{\includegraphics[width=\pwid]{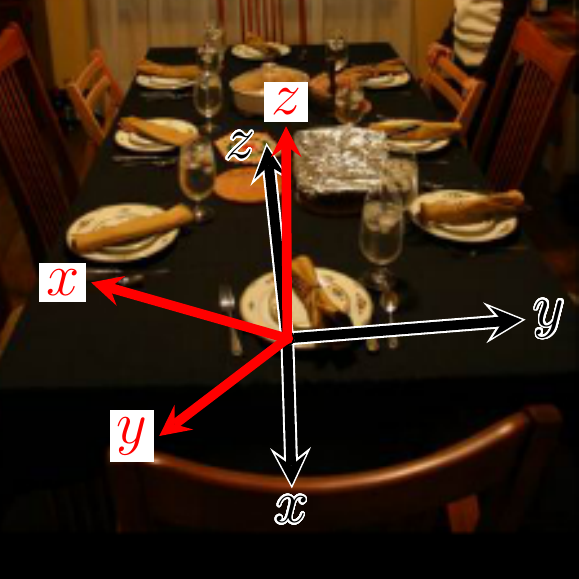}} &
    \multicolumn{1}{c}{\includegraphics[width=\pwid]{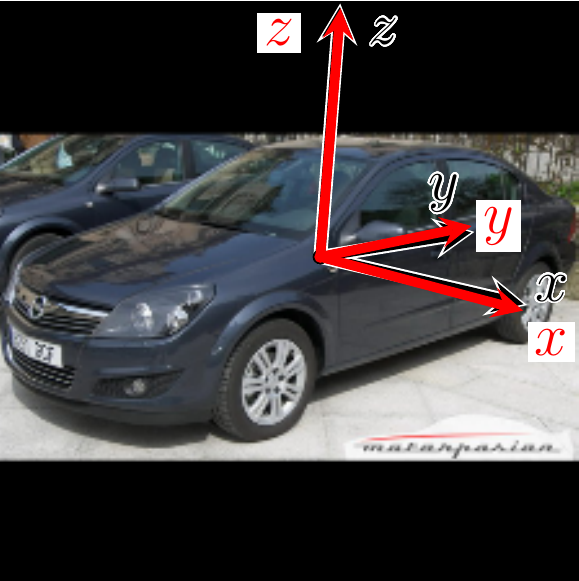}} &
    \multicolumn{1}{c}{\includegraphics[width=\pwid]{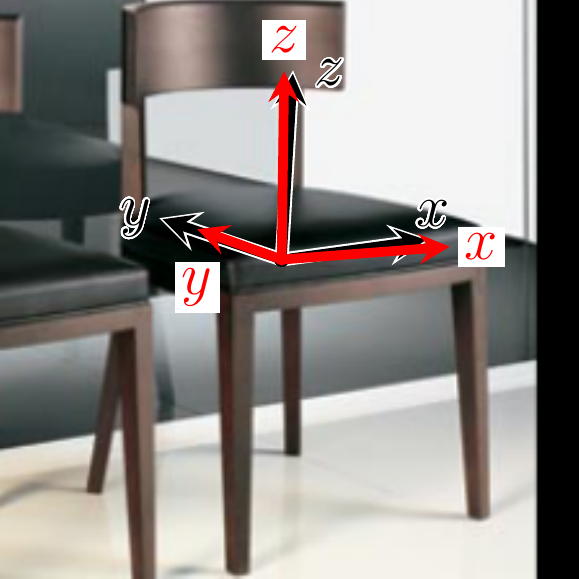}}\\
    \includegraphics[width=\linewidth,trim=170 120 125 60,clip,align=t]{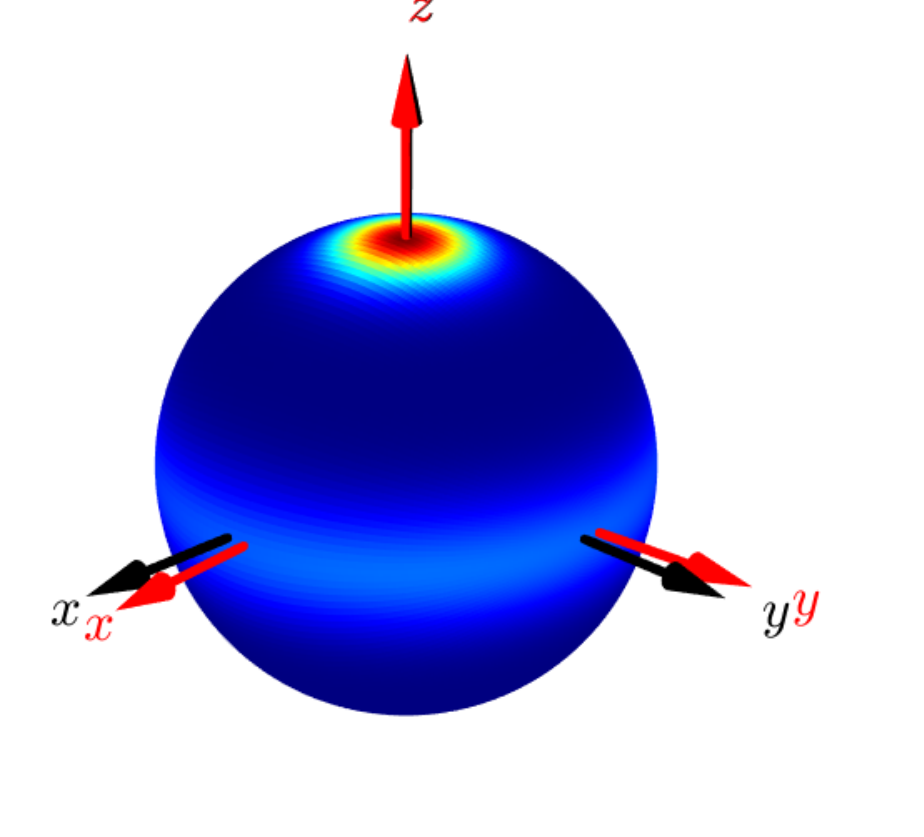} &
    \includegraphics[width=\linewidth,trim=170 120 125 60,clip,align=t]{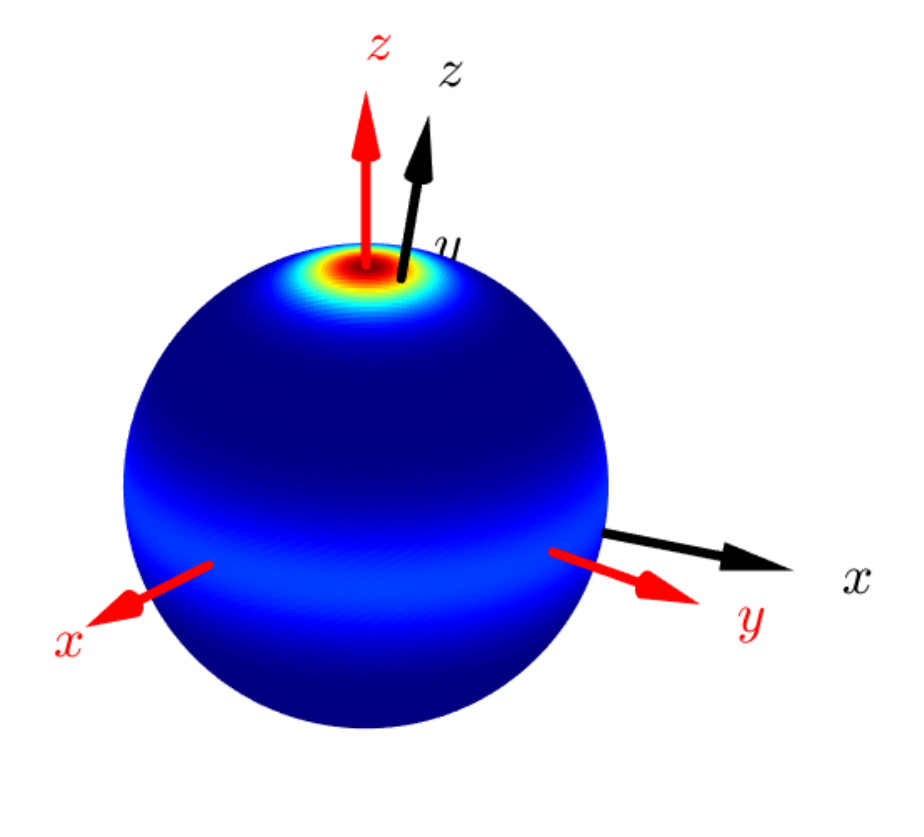} &
    \includegraphics[width=\linewidth,trim=170 120 125 60,clip,align=t]{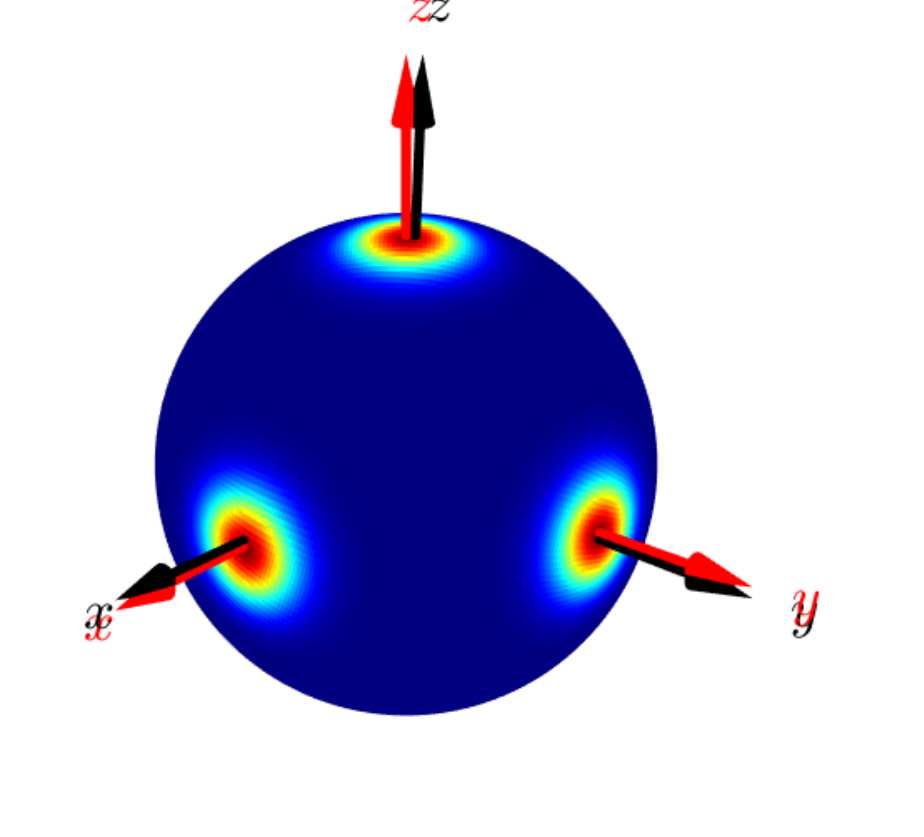} &
    \includegraphics[width=\linewidth,trim=170 80 125 60,clip,align=t]{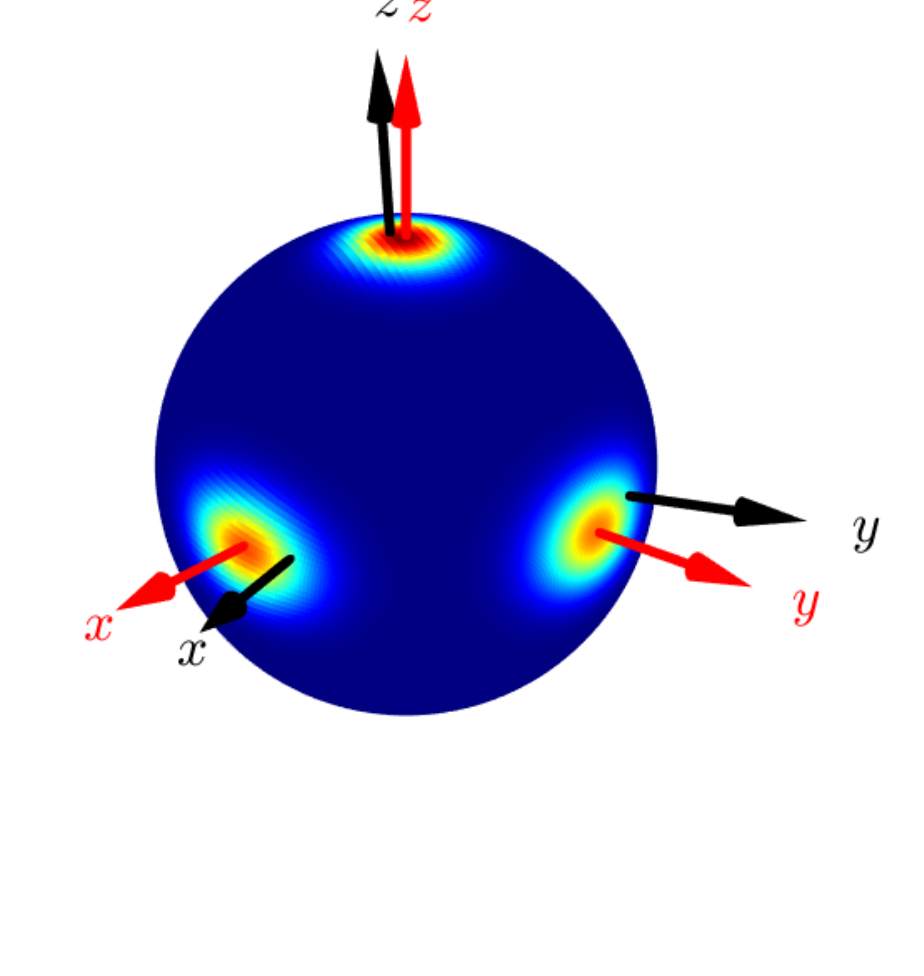}\\[-15pt]
                    \multicolumn{1}{c}{\footnotesize (a)} &
                    \multicolumn{1}{c}{\footnotesize (b)} &
                    \multicolumn{1}{c}{\footnotesize (c)} &
                    \multicolumn{1}{c}{\footnotesize (d)}
  \end{tabular}
  \caption[]{\textbf{Interesting qualitative results on Pascal3D+}. The top row displays example input images with the projected axes displaying the predicted pose (red) and labelled pose (black) of the object. The bottom row shows a visualization of the pdf estimated by the network. The red axis show the maximum likelihood estimate of the rotation matrix estimated from the predicted $F$ matrix, while the axis in black corresponds to the ground truth rotation/pose. For clarity we have aligned the predicted pose with the standard axis. Each probability plot has been scaled independently. The examples shown have been specifically chosen to highlight our algorithm's performance for certain cases: (a)-(b) Examples where model has high uncertainty for the azimuth either due to low resolution or rotation symmetry (c)-(d) Examples where model predicts rotations with high certainty and reasonably low errors.}
  \label{fig:pascal_qual}
\end{figure}

\paragraph{Qualitative results}
Figure \ref{fig:pascal_qual} displays and discusses interesting qualitative results on Pascal3D+ which highlight the probabilistic performance of our method.

\paragraph{Behaviour for classes with rotational symmetries} Several classes in the datasets used have rotational symmetries or effectively have rotational symmetries due to very similar appearance at several distinct viewpoints. Some examples of these classes are canoes, bathtubs, tables, desks, and bottles. Our modelling though is based on a unimodal distribution and here we describe how the model copes with the inherent ambiguity of rotational symmetric objects.

\begin{figure}[t]
  \begin{tabular}{@{}m{.11\linewidth}*{5}{m{.145\linewidth}}}
    \includegraphics[width=\linewidth,trim=20 0 20 0,clip]{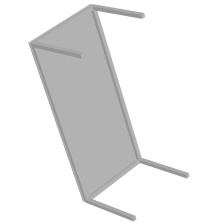} &
    \includegraphics[width=\linewidth,trim=170 105 140 53,clip]{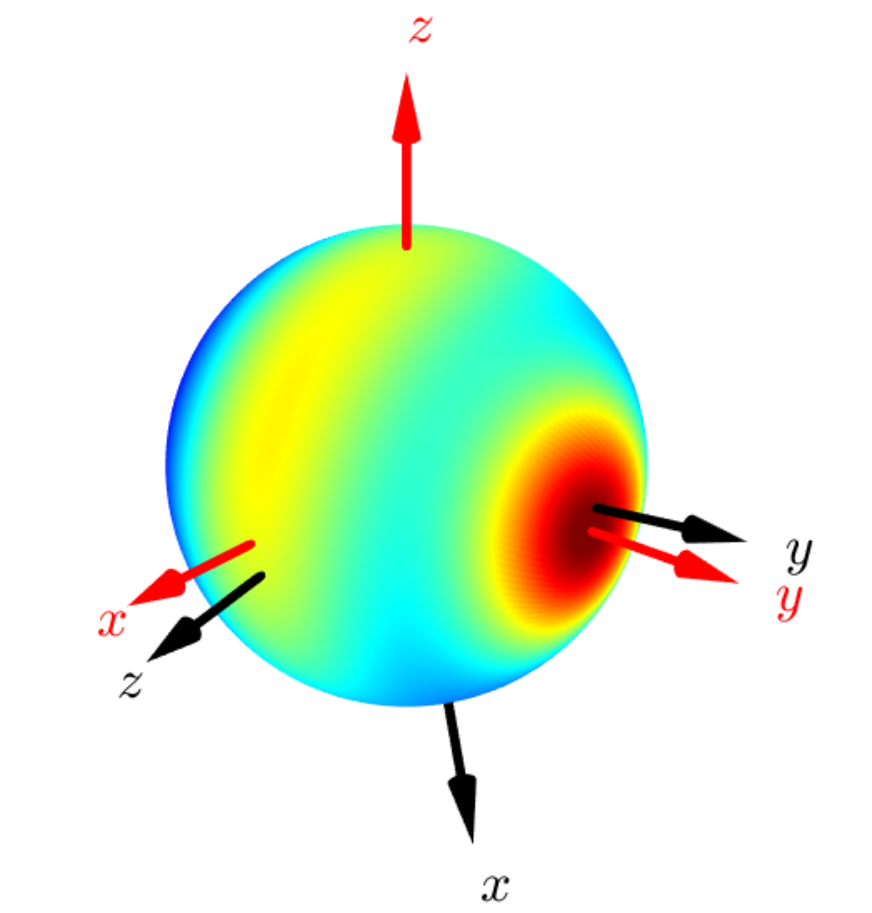} &
    \includegraphics[width=\linewidth,trim=170 105 140 53, clip]{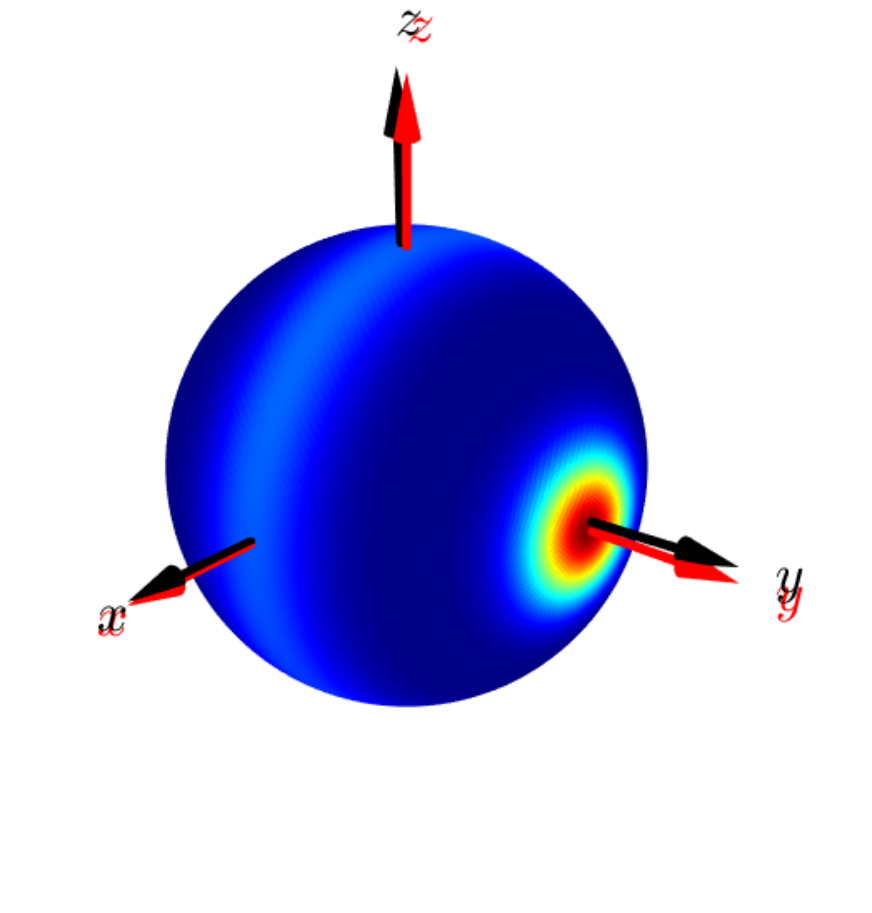} &
    \includegraphics[width=\linewidth,trim=170 105 140 53, clip]{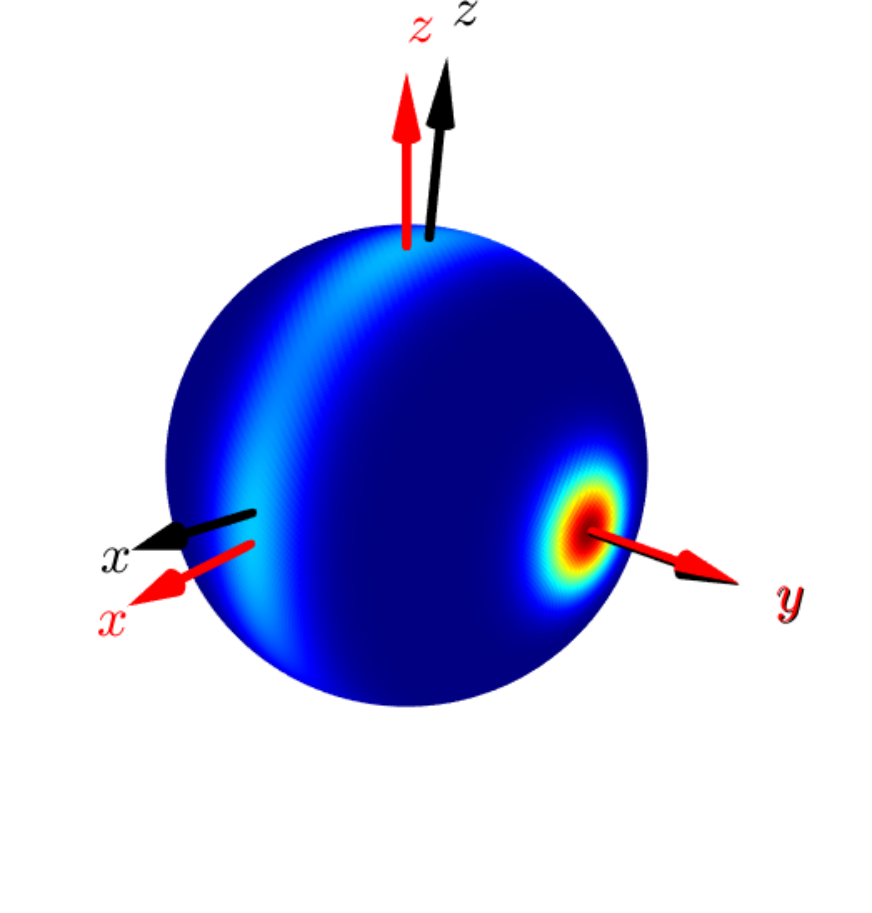} &
    \includegraphics[width=\linewidth,trim=170 105 140 53, clip]{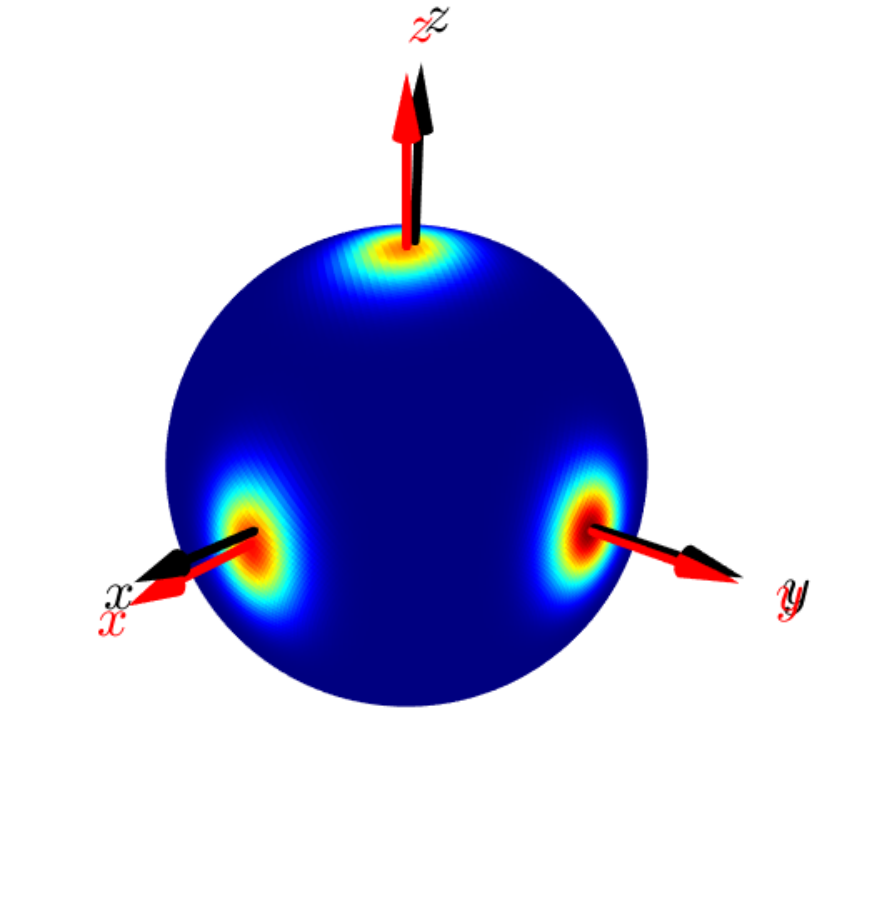}  &
    \includegraphics[width=\linewidth]{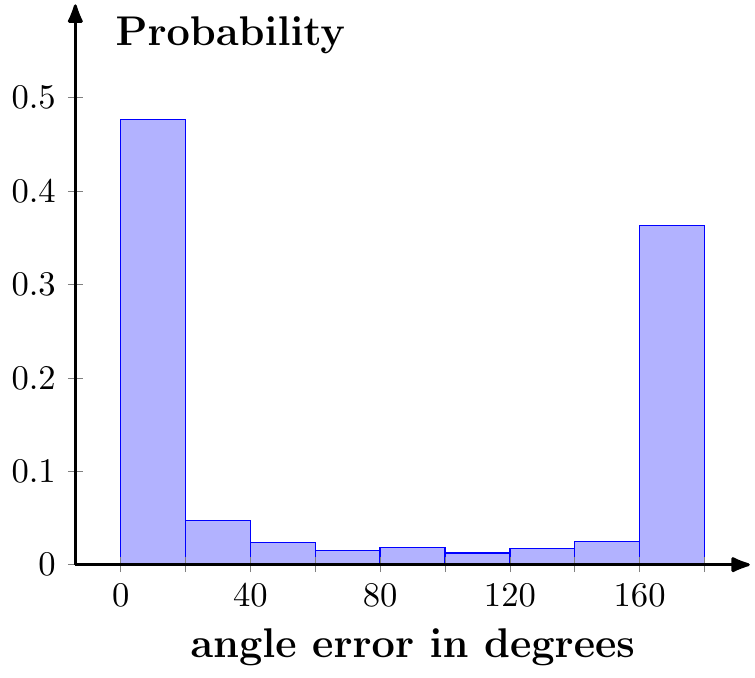} \\
    \mc{ (a) image} & 
    \mc{ (b) $e=1$} &
    \mc{ (c) $e=20$} &
    \mc{ (d) $e=30$} &
    \mc{ (e) $e=40$} &
    \mc{ (f) histogram}
  \end{tabular}
  \caption[]{ \textbf{Evolution of the estimated pdf during training for a rotational symmetric object}. The leftmost figure is a test image from the table class. (b)-(e) Each plot displays the predicted distributions for the object's pose after $e$ epochs of training. The mode of the distribution is shown in red and the ground truth rotation in black. (f) Histogram of the test error angle for the \text{table} class after 50 epochs. See the main text for comments.}
  \label{fig:F_evolution}      
\end{figure}

For Pascal3D+ our method performs well, somewhat surprisingly, for the \textit{dining table} class, see table \ref{tab:pascal_per_class_performance}. However, when the synthetic data is used to augment the training data the performance on this class drops. We suspect the manual labeling process introduces biases for this class with one of the ambiguous poses being labelled much more frequently. But the synthetic data added does not have these biases. This discrepancy between the distribution of training and test set label results in the drop in performance.
For ModelNet10-SO(3) the \textit{table} and \textit{bathtub} classes have rotational symmetries and thus these two classes have much higher median errors than the other classes, see table \ref{tab:modelnet_performance_per_class}. In figure \ref{fig:F_evolution}(f) the histogram of angle errors for the \textit{table} class has a \enquote{U} shape and the median is in the middle of this \enquote{U}. This histogram indicates that at test time the network predicts one of the relevant poses.

To further illuminate this point, plots (b)-(g) of figure \ref{fig:F_evolution} show the evolution of the distribution predicted during training for one specific \textit{table} class test image.  The axis which has no associated ambiguity is identified correctly and confidently early on in training. The other two directions are predicted to have an almost uniform distribution on the plane spanned by the ambiguous axes. This is arguably the best way for the unimodal distribution to describe the situation. In the latter stages of training the network correctly identifies the object's full pose on the training set and uncertainty becomes small. Such behaviour should be considered as a deterioration of the network's probabilistic modelling as it effectively randomly chose one pose from the set of plausible poses and report it is very confident about this decision. Continuing to improve the accuracy on the test set while overfitting the loss often occurs with cross-entropy training of classification networks as well \cite{guo2017calibration}.
The dataset's accuracy and loss plots, in the supplementary material section 2, show our loss is susceptible to this trend too.

\paragraph{Ablation experiments}

We run ablation experiments on Pascal3D+ to identify the importance of the individual components of our approach.  The factors considered are data-augmentation, the class embedding and pre-processing the image via a homography. The results are shown in table \ref{tab:ablation_results}. Rows one and two in table \ref{tab:ablation_results}, show that data augmentation is an important factor. Rows two and three show our warping does not provide a significant improvement by itself on Pascal3D+ compared to cropping. We believe though this warping could be advantageous in many situations and therefore should be used irrespective of these results.
In theory this pre-processing should allow our method to generalize across all pinhole cameras with known intrinsic parameters and negligible radial distortion rather than for cameras with the same intrinsics as Pascal3D+. Rows one and four demonstrate that the class embedding does not give a significant improvement by itself for this dataset. These results further suggest that our loss is the most significant component.

\begin{table}[htpb]
  \renewcommand{\arraystretch}{1.1}
  \caption{ \textbf{Results of ablation experiments on Pascal3D+} for our method. %Specifically we explored the effect of data-augmentation, the pre-processing step of warping the image and the inclusion of the class embedding.
  %The results indicate: 1) Data-augmentation improves performance. 2) Warping to zoom in on the object does not give significant improvement over simply cropping the bounding box. 3) Adding a class embedding does not significantly change the performance.
  }
      \label{tab:ablation_results}  
    \begin{center}
      {
      \begin{tabular}{*{4}{c}*{3}{J{.}{.}{2.1}}}
  \toprule
%  \multicolumn{4}{c}{\textbf{Training regime}}
%  & \multicolumn{3}{c}{\textbf{Performance}}\\
%  \cmidrule(l){1-4}
%  \cmidrule(l){5-7}
  Data aug. & Class embed & Crop & Warp &\mc{MedErr (deg)} & \mc{Acc@${\pi}/ 6$  (\%)} & \mc{Acc@${\pi}/{12}$ (\%)}\\
  \midrule
  \checkmark & \checkmark & $\times$ & \checkmark & \multicolumn{1}{B{.}{.}{2,1}}{8.9}  & 90.8 & \multicolumn{1}{B{.}{.}{2,1}}{74.5}\\
  $\times$ & \checkmark & $\times$ & \checkmark & 10.1 & 88.4 & 69.5\\
  $\times$ & \checkmark & \checkmark & $\times$  & 10.0 & 88.4 & 70.0\\
  \checkmark & $\times$ & $\times$ & \checkmark  & \multicolumn{1}{B{.}{.}{2,1}}{8.9} & \multicolumn{1}{B{.}{.}{2,1}}{90.9} & 74.1\\
  \bottomrule
  \end{tabular}
  }
  \end{center}
\end{table}

%% file: SupplementaryMaterial.tex
\section {Additional experimental results}
\subsection{Additional quantitative results}

In table \ref{tab:pascal_per_class_performance_all} We show the performance for our method across all classes in Pascal3D+

\begin{table}[t]
\caption{\textbf{Pascal3D+ per-class performance} of our method, w/o and with synthetic training data, compared to the state-of-the-art method \citet{mahendran2018mixed} on this dataset. The top three rows report the median angle error per class measured in degrees. The bottom three rows report Acc@${\pi}/{6}$ measured as a percentage.}
\label{tab:pascal_per_class_performance_all}  
\centering
\addtolength{\leftskip} {-4cm}
\addtolength{\rightskip}{-4cm}
\begin{tabular}{l*{13}{J{.}{.}{2.1}}@{}}
\toprule
\textbf{Method} & \mc{aero} & \mc{bike} & \mc{boat} & \mc{bottle} &  \mc{bus} & \mc{car} & \mc{chair} & \mc{dtable} & \mc{mbike} & \mc{sofa} & \mc{train} & \mc{tv} & \mc{\textbf{mean}} \\
\midrule
\citep{mahendran2018mixed} & 8.5 & 14.8 &
20.5 & \multicolumn{1}{B{.}{.}{2,1}}{7.0} & \multicolumn{1}{B{.}{.}{2,1}}{3.1} &
5.1 & 9.3 & 11.3 & 14.2 &
10.2 & 5.6 & 11.7 & 10.1 \\
Ours w/o & 10.2 & 14.7 & 12.6 & 8.2 & 3.5 & \multicolumn{1}{B{.}{.}{2,1}}{3.8} & 7.8 & \multicolumn{1}{B{.}{.}{2,1}}{8.4} & 13.1 & 7.7 & 5.4 & 11.4 & 8.9\\
Ours with & \multicolumn{1}{B{.}{.}{2,1}}{6.8} & \multicolumn{1}{B{.}{.}{2,1}}{12.3} & \multicolumn{1}{B{.}{.}{2,1}}{12.2} & 7.6 & 3.8 & 3.9 & \multicolumn{1}{B{.}{.}{2,1}}{6.4} & 10.6 & \multicolumn{1}{B{.}{.}{2,1}}{10.8} & \multicolumn{1}{B{.}{.}{2,1}}{7.4} & \multicolumn{1}{B{.}{.}{2,1}}{5.3} & \multicolumn{1}{B{.}{.}{2,1}}{10.3} & \multicolumn{1}{B{.}{.}{2,1}@{}}{8.1}\\
\midrule
\citep{mahendran2018mixed}  & 87.0 & 81.0 & 64.0 & \multicolumn{1}{B{.}{.}{2,1}}{96.0} & 97.0 & 95.0 & 92.0 & 67.0 & 85.0 & 97.0 & 82.0 & 88.0 & 85.9 \\
Ours w/o & 87.9 & 82.3 & 77.8 & 95.5 & 98.4 & 98.3 & 94.8 & \multicolumn{1}{B{.}{.}{2,1}}{82.0} & 88.0 & 98.1 & 97.9 & 89.0 & 90.8 \\
Ours with & \multicolumn{1}{B{.}{.}{2,1}}{93.9} & \multicolumn{1}{B{.}{.}{2,1}}{89.2} & \multicolumn{1}{B{.}{.}{2,1}}{80.0} & 94.6 & \multicolumn{1}{B{.}{.}{2,1}}{98.6} & \multicolumn{1}{B{.}{.}{2,1}}{99.1} & \multicolumn{1}{B{.}{.}{2,1}}{98.7} & 78.0 & \multicolumn{1}{B{.}{.}{2,1}}{93.1} & \multicolumn{1}{B{.}{.}{2,1}}{99.7} & \multicolumn{1}{B{.}{.}{2,1}}{98.4} & \multicolumn{1}{B{.}{.}{2,1}}{94.2} & \multicolumn{1}{B{.}{.}{2,1}@{}}{93.1} \\
  \bottomrule
\end{tabular}  
\end{table}

\subsection{Additional qualitative results}

\begin{figure}[!b]
    \centering
    \includegraphics[width=\linewidth]{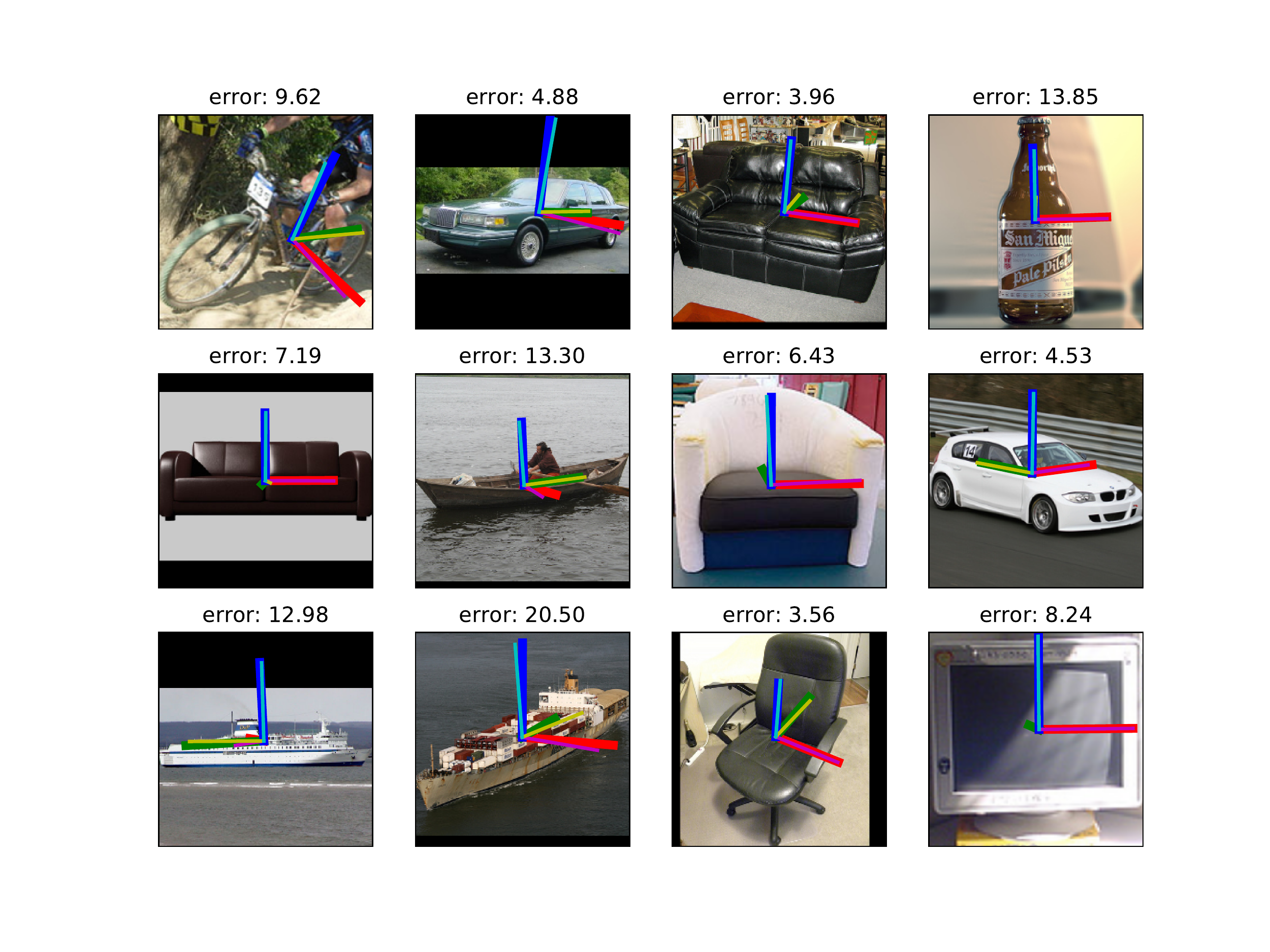}
    \caption{\textbf{Pascal3D+ predictions.} Visualization of a random subset of the test set for Pascal3D+. The maximum likelihood rotation for the probability distribution predicted by the network (thick lines) compared to the ground truth (thin lines). Background image is the preprocessed input}
    \label{fig:random_errors}
\end{figure}

Figure \ref{fig:random_errors} show how well our method works on a random subset of the Pascal3D+ test set.

\section{Overfitting confidences}
\begin{figure}[t]
    \def\pwid{.5\linewidth}
    \centering
    \includegraphics[width=\pwid]{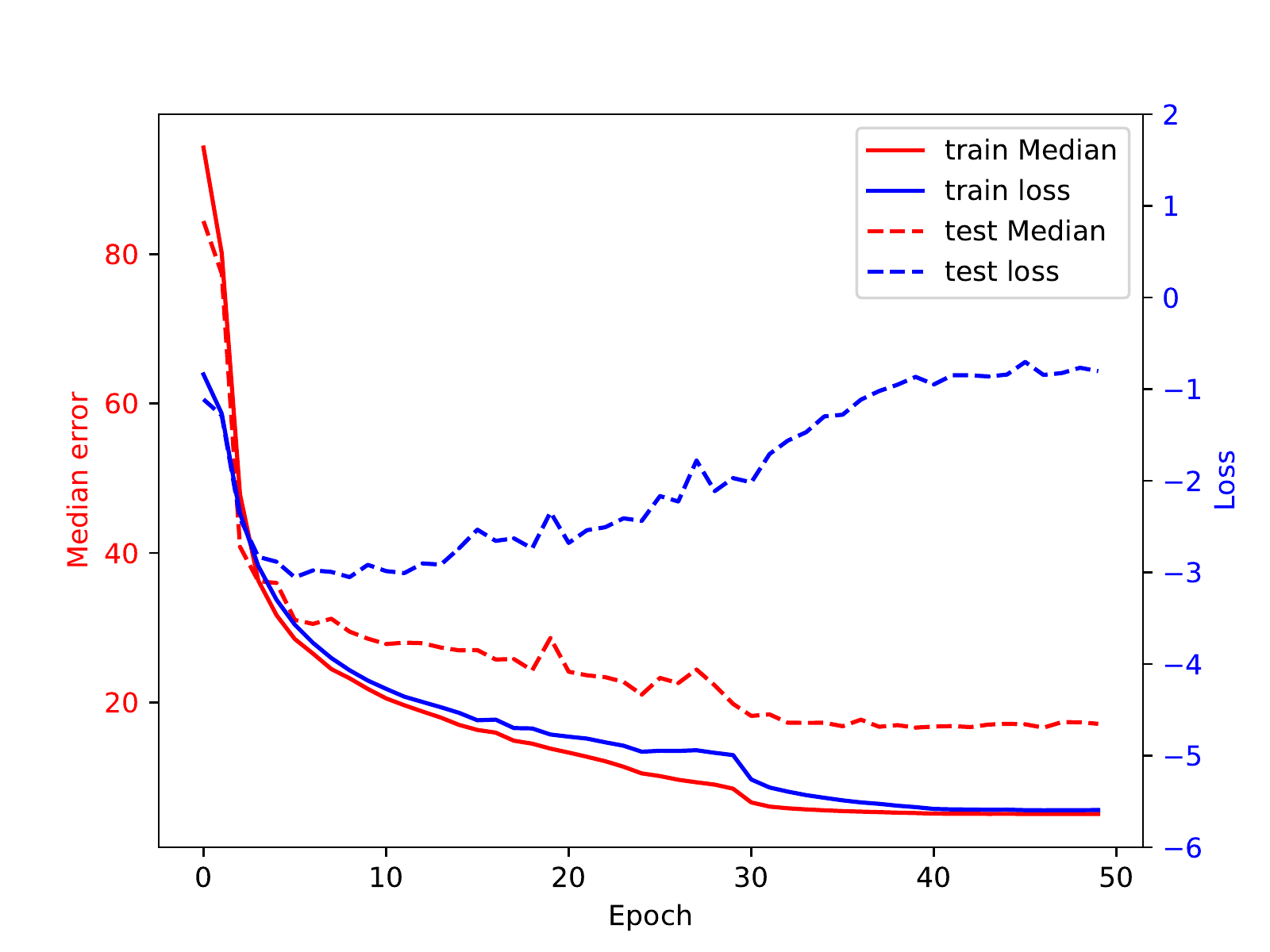}
    \caption{\textbf{Median error compared to mean loss} Visualization of how the median error and median loss changes over time when training on ModelNet10-SO(3)}
    \label{fig:loss_vs_acc}
\end{figure}

In figure \ref{fig:loss_vs_acc} we see how we overfit more on the loss than the median error. A similar effect have been observed when training networks to perform classification \cite{guo2017calibration}.
This effect was more pronounced for ModelNet10-SO(3) due to the prominence of rotation ambiguous samples.

\section{Geometric interpretation of the matrix Fisher distribution}
\label{sec:geo_von_mises}

To further help understanding how the shape of the distribution relates to $F$, \cite{lee2018bayesian} defines the \textit{proper SVD} of $F$ as:
{\small
  \begin{align}
    %% F = U_1 S' V_1^T = \underset{U}{\underbrace{U_1 \, \text{diag}(1, 1, \text{det}(U_1))}}\; \underset{S}{\underbrace{\text{diag}(s'_1, s'_2, \text{det}(U_1 V_1)s'_3)}}
    %% \, \underset{V^T}{\underbrace{\text{diag}(1, 1, \text{det}(V_1)) V_1^T}} = U S V^T
  F = U_1 S' V_1^T
    &= \underset{U}{\underbrace{U_1  \begin{bmatrix}
      1 & 0 & 0 \\
      0 & 1 & 0 \\
      0 & 0 & \text{det}(U_1)
    \end{bmatrix}}}
    \underset{S}{\underbrace{\begin{bmatrix}
      s'_1 & 0 & 0 \\
      0 & s'_2 & 0 \\
      0 & 0 & \text{det}(U_1\,V_1) s'_3
    \end{bmatrix}}}
    \underset{V^T}{\underbrace{\begin{bmatrix}
      1 & 0 & 0 \\
      0 & 1 & 0 \\
      0 & 0 & \text{det}(V_1)
    \end{bmatrix}
        V_1^T}}
    = U S V^T %\nonumber
\end{align}}\normalsize
so that $U$ and $V$ are guaranteed to be rotation matrices and $S$ contains the \textit{proper singular values} of $F$ with $s_1 \geq s_2 \geq |s_3|$. Note that $\hat{R} = U V^T$. The columns of $U$ define three orthogonal axes directions around which the mode rotation can be rotated. As the angle of a rotation about the axes, $U e_i$, varies from 0 to 360 the peakedness of the pdf along the great circle, traced out by $\hat{R} e_j$ or $\hat{R} e_k$, is proportional to $(s_j + s_k)$ with $j, k \in \{1, 2, 3\}\backslash i$. So if $(s_j + s_k)$ is small then the distribution along the great circle approximates a uniform distribution while if it is large then the distribution is very peaked. The columns of $U$ are termed the \textit{principal axes} and can be thought of as analogous to the \textit{principal axes} of a multivariate Gaussian. Figure \ref{fig:von_mises_spread2} gives a visualization of examples of this interpretation.

\def\pwid{.18\linewidth}
\begin{figure}[t]
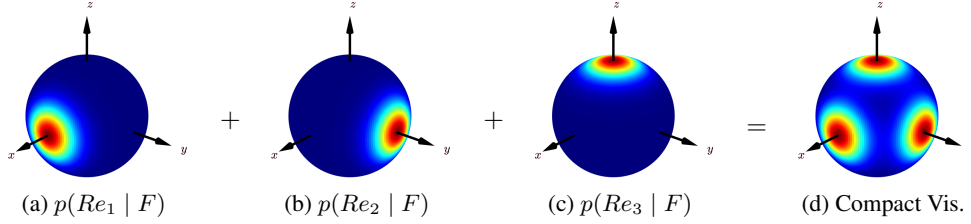

  \begin{tabular}{*{4}{m{\pwid}m{.01\linewidth}}}
    \includegraphics[width=\linewidth,trim=170 120 140 60,clip]{images/VonMisesExamples/von_mises_1_1.pdf} &
   {\vspace*{20pt}$+$} &
    \includegraphics[width=\linewidth,trim=170 120 140 60,clip]{images/VonMisesExamples/von_mises_1_2.pdf} &
    {\vspace*{20pt}$+$} &
    \includegraphics[width=\linewidth,trim=170 120 140 60,clip]{images/VonMisesExamples/von_mises_1_3.pdf} &
    {\vspace*{25pt}$=$} &
    \includegraphics[width=\linewidth,trim=170 120 140 60,clip]{images/VonMisesExamples/von_mises_1_all.pdf}\\
                    \multicolumn{1}{c}{\footnotesize (a) $p(R e_1 \mid F)$} & &
                    \multicolumn{1}{c}{\footnotesize (b) $p(R e_2 \mid F)$} & &
                    \multicolumn{1}{c}{\footnotesize (c) $p(R e_3 \mid F)$} & &
                    \multicolumn{1}{c}{\footnotesize (d) Compact Vis.} 
                    
  \end{tabular}
  \caption[]{\textbf{Visualizing the matrix Fisher distribution on $SO(3)$.} Visualization of the probabilities for the case $F=\text{diag}(5, 5, 5)$. Let $e_1, e_2$ and $e_3$ correspond to the standard basis of $\mathbb{R}^3$ and is shown by the black axes. (a) This plot shows the probability distribution of $R e_1$ when $R \sim \mathcal{M}(F)$. Thus the pdf shown on the sphere corresponds to the probability of where the $x$-axis will be transformed to after applying $R \sim \mathcal{M}(F)$. (b) and (c) Same comment as (a) except consider $e_2$ and $e_3$ instead of $e_1$. (d) A compact visualization of the plots in (a), (b) and (c) is obtained by summing the three marginal distributions and displaying them on the 3D sphere. All four plots are plotted within the same scale and a \textit{jet} colormap is used.}
  \label{fig:von_mises_vis}
\end{figure}

\begin{figure}[t!]
  \centering
  \def\phei{.1\textheight}
  \begin{tabular}{cccccc}
    \includegraphics[height=\phei,trim=170 120 140 60,clip]{images/VonMisesExamples/von_mises_2_all.pdf} &
    \includegraphics[height=\phei,trim=170 120 140 60,clip]{images/VonMisesExamples/von_mises_3_all.pdf} &
    \includegraphics[height=\phei,trim=130 120 140 60,clip]{images/VonMisesExamples/von_mises_4_all.pdf} \\
                    {\small (a) $\text{diag}(20, 20, 20)$} &
                    {\small (b) $\text{diag}(25, 5, 1)$} &
                    {\small (c) $A_3\, \text{diag}(25, 5, 1)$}\\[3pt]    
    \includegraphics[height=\phei,trim=170 120 140 60,clip]{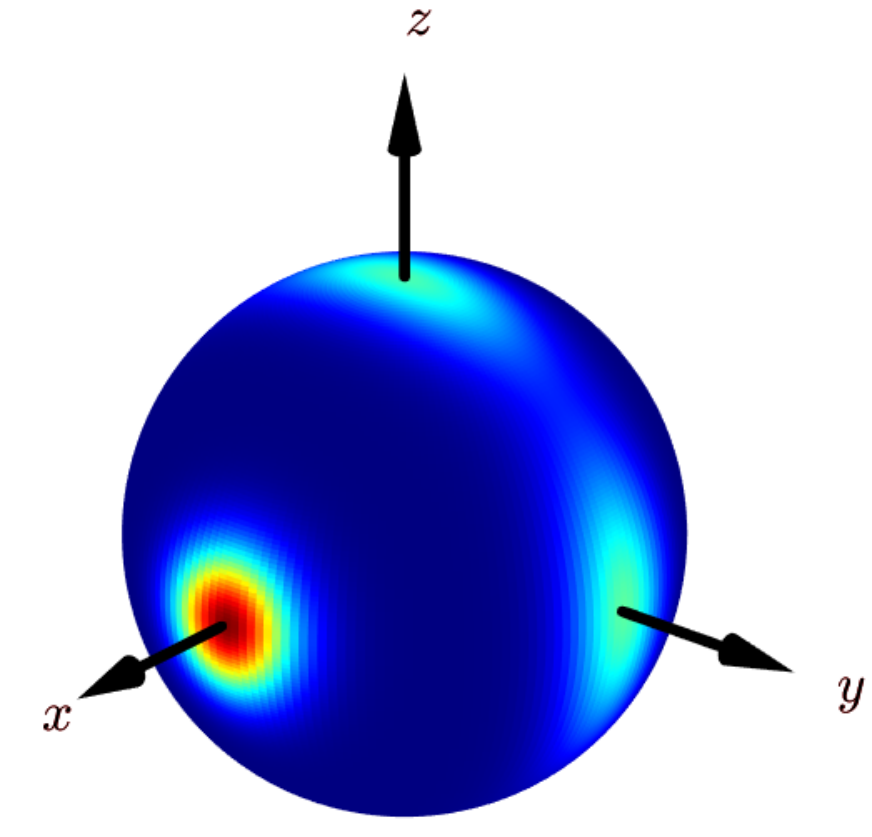} &
    \includegraphics[height=\phei,trim=170 120 140 60,clip]{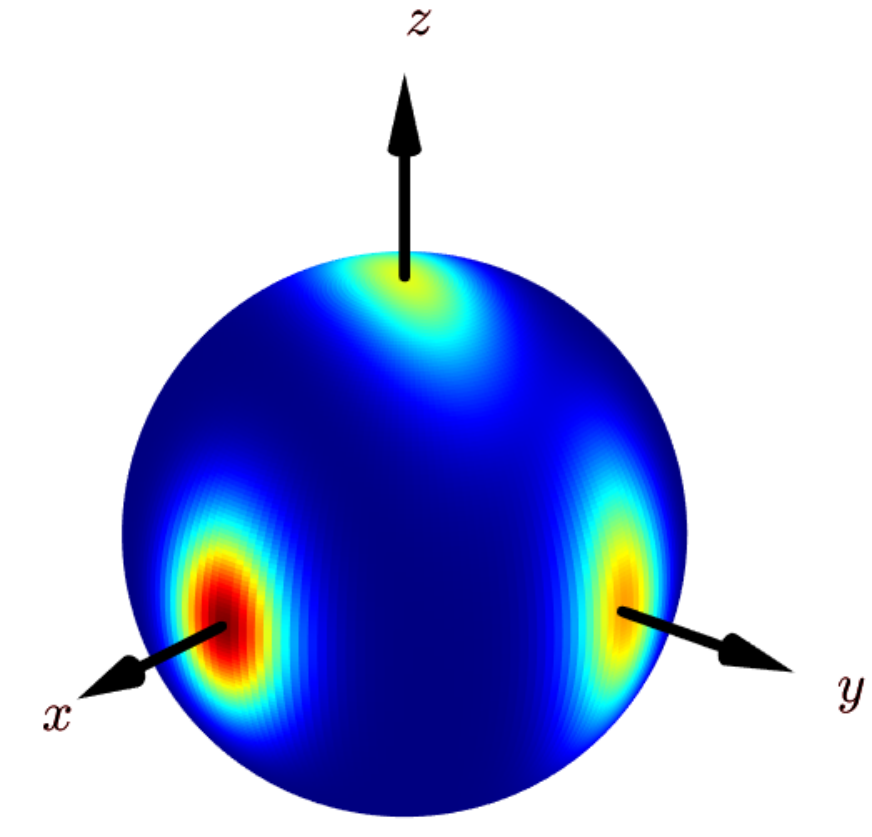} &
    \includegraphics[height=\phei,trim=170 120 140 60,clip]{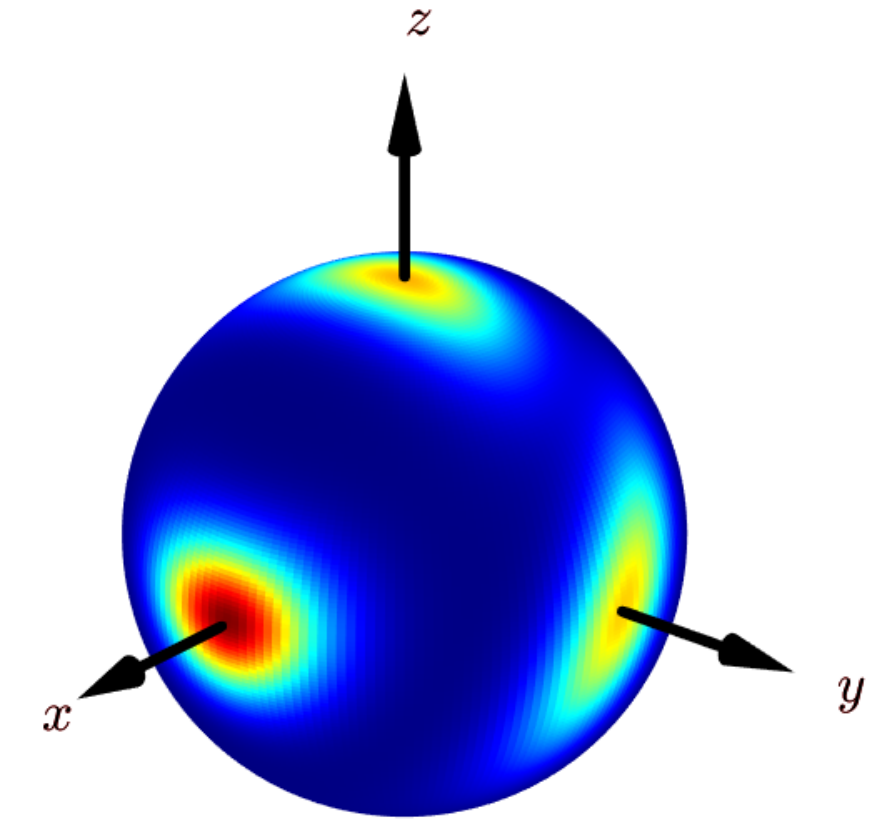} \\
                    %{\small $F = \text{diag}(5, 5, 5)$} &
                    {\small (d) $A_1\, \text{diag}(25, 5, 1) A_1^T$} &
                    {\small (e) $A_2\, \text{diag}(25, 5, 1) A_2^T$} &
                    {\small (f) $A_3\, \text{diag}(25, 5, 1) A_3^T$}
  \end{tabular}
  \caption[]{\small \textbf{Effect of $F$ on shape of the matrix Fisher distribution}. Below each plot is the value of $F$. Each $A_i$ corresponds to the rotation matrix obtained by rotating by $-\pi/6$ degrees around $e_i$. (a) For a spherical $F$ the mode of the distribution is the identity and the principal axes correspond to $e_1, e_2, e_3$. The distribution for each axis is circular and identical and more peaked than in figure \ref{fig:von_mises_vis} as the singular values are larger. (b) Have a diagonal $F$ and thus the mode and the principal axes once again coincide. The distributions for the $y$- and $z$-axes are more elongated than for the $x$-axis as the first singular value dominates. (c) Here $F$ is obtained by pre-multiplying a diagonal matrix by $A_3$. Thus the mode corresponds to $A_3$ and is shown by the red axes. The shape of the distributions for each axis though remain the same as in (b). (d, e, f) In each of these plots the diagonal matrix from (b) is pre-multiplied by $A_i$ and post-multiplied by $A_i^T$. Thus the mode rotation is the identity. However, the principal axes correspond to the columns of $A_i$. Thus the axis distributions are centred at the standard location but the orientation of the spread has been affected by the direction of the principal axes.}
  \label{fig:von_mises_spread2}
\end{figure}

\newpage
\section{Properties of loss}

For more convenient notation we will introduce a flattening function $h: \mathbb{R}^{m\times m} \rightarrow \mathbb{R}^{m^2}$ s.t. $h(x)_{(i-1)*m+j} = x_{i,j} \forall i,j \in \{1,2, \cdots m\}$ and an inflation function $g$ such that $g = h^{-1}$.

In this section F is an element of $\mathbb{R}^9$. $||.||_F$ is the frobenius norm. $||.||_2$ for vectors is the traditional $L_2$ norm. At one place we use the matrix 2 norm which is the magnitude of the largest eigenvalue. We will use the nonstandard notation $||.||_{M2}$ for this norm to avoid confusion. The standard notation for this norm is $||.||_{2}$.
We will use the fact that $tr(g(F)^TR) = F^Th(R)$ and $||g(F)||_F = ||F||_2$

\subsection{Lipschitz continous}
Here we show that the loss is $\alpha$-Lipschitz continous for $\alpha$=6. This is equivalent to the $L_2$ norm of the gradient being less than 6.
\begin{equation}
    Loss(F, R) = log (a(g(F)))-tr(g(F)^TR)
\end{equation}
The gradient is
\begin{equation}
    ||\nabla_F Loss(F, R)||_2 = ||\nabla_F log(a(g(F))) - h(R)||_2 \le ||\nabla_F log(a(g(F)))||_2 + ||-h(R)||_2
\end{equation}

The last step follows from triangle inequality

We know $||h(R)||_2$ = $||R||_F$ = 3 since it is a rotation matrix.

We now use the definition of a(F) to compute the gradient

\begin{gather}
    \nabla_F log(a(g(F))) = \dfrac {\nabla_F a(g(F))} {a(g(F))} = \dfrac {1} {a(g(F))} \nabla_F \int_{h(R) \in SO(3)} \exp(tr(g(F)^TR)) dR = \\
    \dfrac {1} {a(g(F))} \int_{R \in SO(3)} h(R) \exp(F^Th(R))) dR = E[h(R)|F]
\end{gather}

Due to convexity of frobenius norm and Jensen's inequality we have $||\nabla_F log(a(g(F))) ||_2 = ||E[h(R)|F])||_2= ||E[R|F]||_F \le E[||R||_F|F] = 3$
This concludes the proof.

\subsection{Convexity}

We first compute the hessian of $log(a(g(F)))$

We already have $\nabla_F log(a(g(F))) = \dfrac {1} {a(g(F))} \int_{R \in SO(3)} h(R) \exp(tr(g(F)^TR)) dR$ from previous section.

We differentiate again to get
\begin{gather}
    (\nabla^2 log(a(g(F))))_{i,j} =\dfrac 1 {a(g(F))} \int_{R \in SO(3)} h(R)_i h(R)_j \exp(tr(g(F)^TR)) dR - \\
    \dfrac 1 {(a(g(F)))^2} \int_{R \in SO(3)} h(R)_i \exp(tr(g(F)^TR)) dR \int_{R \in SO(3)} h(R)_j \exp(tr(g(F)^TR)) dR = \\
    E[h(R) h(R)^T]_{i,j} - (E[h(R)] E[h(R)]^T)_{i,j} = Var[h(R)]_{i,j}
\end{gather}
Since a variance matrix is positive semidefinite it follows that the hessian is as well. Therefore this term is convex
The term $tr(g(F)^TR)$ is linear. Linear functions are convex.
The set of convex functions are closed under addition. Therefore the loss is convex with respect to the network output $F$.

\subsection{Lipschitz continous gradients}
A function has $\beta$-Lipschitz continous gradients if the largest eigenvalue of the hessian is less than $\beta$.

We know
\begin{gather}
    ||\nabla^2 log(a(g(F))))||_{M2} = ||Var[h(R)]||_{M2} \le ||Var[h(R)]||_F = \\ ||E[h(R)h(R)^T]-E[h(R)]E[h(R)]^T||_F \le ||E[h(R)h(R)^T]||_F \le \\ E[||h(R)h(R)^T||_F] \le E[||h(R)||_F^2] = 9
\end{gather}

Therefore this function has $\beta$-Lipschitz continous gradients with $\beta$=9

\section{Approximating the normalizing constant}
\label{sec:approx_norm_constant}

\begin{figure}[htpb]
    \centering
    \includegraphics[scale=0.45]{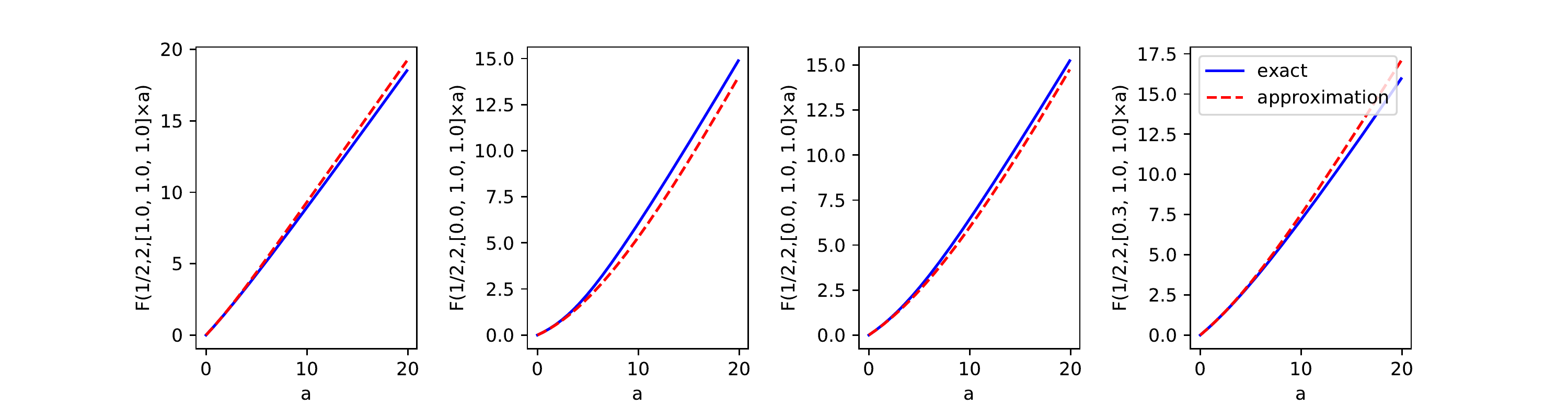}
    \caption{Illustration of approximation for the forward pass, blue line is true function computed by software from \cite{koev2006efficient}, red dashed is our approximation.
    plot visualizing one dimensional slices as the magnitude of the norm of the argument increases.
    }
    \label{fig:hyper_forward}
\end{figure}

\begin{figure}[htpb]
    \centering
    \includegraphics[scale=0.45]{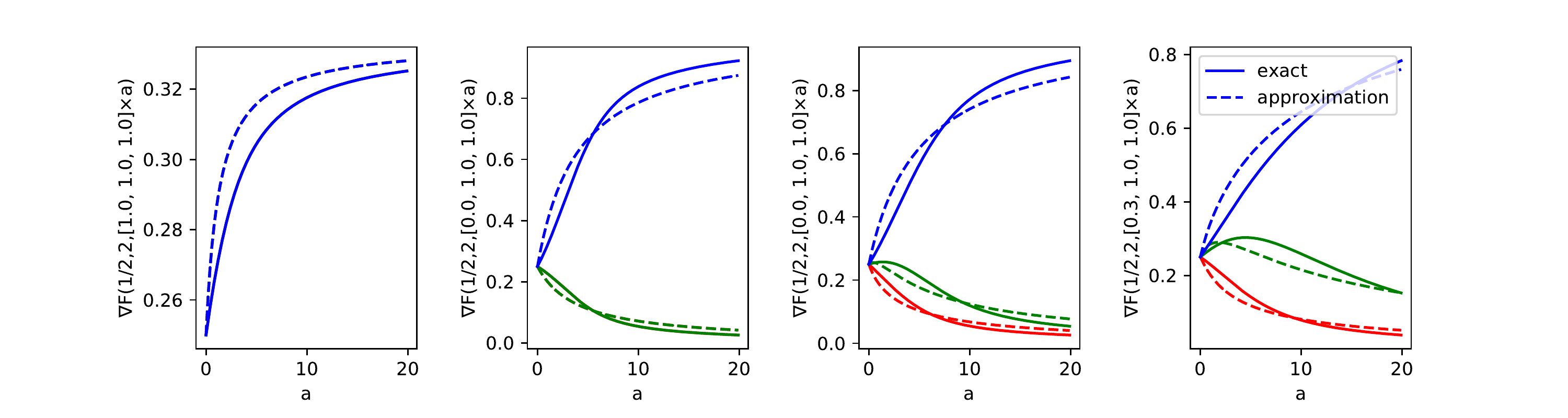}
    \caption{Illustration of the gradient of the hypergeometric function. Colors correspond to the three different input variables. Solid line is numerical differentiation of \cite{koev2006efficient} while dashed line is our approximation.
    }
    \label{fig:hyper_diff}
\end{figure}

The generalized hypergeometric function of matrix arguments can be defined recursively by integrals over positive definite matrices \cite{herz1955bessel}. Similar to the standard generalized hypergeometric function it has a combinatorial definition as well which is
\begin{gather}
{_1}F_1^{(2)}(\tfrac{1}{2}, 2, X) \sum\limits_{k=0}^{\infty} \sum\limits_{\kappa \vdash k} \dfrac {(\tfrac {1} {2})_{\kappa}^{(2)}} {k! {(2)}^{(2)}_{\kappa}} C_\kappa^{(2)}(X)
\end{gather}

For details see \cite{koev2006efficient}

The generalized hypergeometric function of matrix argument can be computed for example by the method used in \cite{koev2006efficient}. But this method becomes slow or unstable when the singular values become large. Another way to approximate the function is by numerical integration over a sphere, but this method is also slow. To rectify this we identify by visual inspection of sampled values that $\log({}_1F_1(\tfrac{1}{2}, 2, [x*a,y*a,z*a,0]))$ behaves asymptotically like $a \max(x,y,z)$ for large $a$. We then use moment matching at the origin to fit a function of the form.

\begin{align}
  \log\left({}_1F_1(\tfrac{1}{2}, 2, [x,y,z])\right) &\approx h([x,y,z])\\
  &= \max(x,y,z) + f_1(x,y,z)\, (1-\exp(f_2(x,y,z)\,\max(x,y,z))
\end{align}
where $f_1$ and $f_2$ are second degree polynomials chosen in such a way that the gradient and hessian match the function we want to approximate at the origin.

To get better approximations of the gradient we notice from the integral, if all 4 diagonal elements of $\Lambda$ are increased by $\epsilon$ the integral will be scaled by $\exp(\epsilon)$. Therefore
\begin{align}
 \sum\limits_{i = 1}^4 \dfrac {d} {d x_i}  \log(_1F_1(\tfrac {1} {2}, 2, [x_1,x_2,x_3,x_4])) = 1 
\end{align}

By using the software from \cite{koev2006efficient} to compute the value of this function and its numerical gradients for a few points accurately and then fitting a simplified function to these values to get a more accurate estimate of the gradient

\begin{gather}
    \nabla_{\Lambda} \log\left({}_1F_1(\tfrac{1}{2}, 2, \Lambda)\right)
\approx
    \dfrac
      {1 + \Lambda}
      {\sum\limits_{i=1}^{4} 1 + \Lambda_i}
\end{gather}
Denominator comes from having to fulfill equation 25.

\section{Deriving normalizing constant of the Fisher distribution} \label{normalizing_factor}
\subsection{Scale factor for changing between rotation matrices and quaternions}
\label{scale_change_quat_to_rot}
Consider the two quaternions $q_1 = a + b\,\textbf{i}+c\,\textbf{j}+d\,\textbf{k}$ and $q_2 = e + f\,\textbf{i}+g\,\textbf{j}+h\,\textbf{k}$. Then
\begin{align}
q_1 q_2 = \begin{bmatrix}
    a & -b & -c & -d \\
    b & a  & -d & c  \\
    c & d  & a  & -b \\
    d & -c & b  & a  \\
\end{bmatrix} \ \begin{bmatrix}
    e \\
    f \\
    g \\
    h \\
\end{bmatrix}
\end{align}
As we can see multiplication of a unit quaternion with another quaternion is isometric since this matrix is orthonormal.
We know the Frobenius norm does not change with respect to rotations either. From this we can conclude that the scale change when doing a basis change from rotation matrices to quaternions is constant.

\subsection{Deriving an expression for the normalizing constant}
\label{sec:NormFact}

The normalizing constant, $a(F)$ for the Fisher distribution with parameter matrix $F$ is found by calculating this integral:
\begin{equation}
  a(F) = \int\limits_{R \in SO(3)} e^{\text{tr}(F^TR)} dR
\label{eqn:normfactor}  
\end{equation} 
where $dR$ is the measure induced by the Frobenius norm.

We can decompose $F = U^T \Lambda V$. $U$ and $V$ are chosen to be right handed and if any singular value is negative it is the one with the smallest absolute value. Denote the singular values in descending order as $s_1, s_2, s_3$. From the SVD of $F$ we can write the exponent in equation (\ref{eqn:normfactor}) as  
\begin{align}
  \text{tr}(F^TR) = \text{tr}(V^T \Lambda U R) = \text{tr}(\Lambda URV^T) = \text{tr}(\Lambda \hat{R})
\end{align}
where $\hat{R} = URV^T$. Thus equation (\ref{eqn:normfactor}) becomes after a basis change
\begin{gather}
    a(F) = \int\limits_{R \in SO(3)} e^{tr(F^TR)} dR =
    %% \int\limits_{R \in SO(3)} e^{tr(V^TSUR)} dR = \\
    %% \int\limits_{R \in SO(3)} e^{tr(SURV^T)} dR = \\
    \int\limits_{\hat{R} \in SO(3)} e^{tr(\Lambda \hat{R})} d\hat{R}
\end{gather}
No basis change scale factor is needed since the measure is rotation invariant.

Denote $R_q: S(3) \rightarrow SO(3)$ as the mapping from a quaternion to the corresponding rotation matrix and let $\Lambda_1$ and $\Lambda_2$ be $4 \times 4$ matrices defined as:
\begin{align}
  \Lambda_1 &= \text{diag}(s_1-s_2-s_3, s_2-s_1-s_3, s_3-s_1-s_2, s_1+s_2+s_3)\\
  \Lambda_2 &= \text{diag}(2(s_1-s_3), 2(s_2-s_3), 0, 2(s_1+s_2))
\end{align}
Let $\mathbf{q} = (x, y, w, z)^T$ be a quaternion then
\begin{align}
  \text{tr}\left(\Lambda R_q(\mathbf{q})\right) &= s_1(x^2 + w^2 - y^2 - z^2) + s_2(y^2 + w^2 - x^2 - z^2) + s_3(z^2 + w^2-x^2-y^2)\\
  &= \mathbf{q}^T \Lambda_1 \mathbf{q}
  \label{eqn:intermediate_result1}
\end{align}
Using this result and the newly introduced notation we can further tackle solving the integral in equation (\ref{eqn:normfactor}  )

\begin{align}
  \int\limits_{\hat{R} \in SO(3)} e^{\text{tr}\left(\Lambda\hat{R}\right)} d\hat{R} \,=\,
   C \int\limits_{\mathbf{q} \in S(3)} e^{\text{tr}\left(\Lambda R_q(\mathbf{q})\right)} d\mathbf{q}
  &= C \int\limits_{\mathbf{q} \in S(3)} e^{\left(\mathbf{q}^T \Lambda_1 \mathbf{q}\right)} d\mathbf{q}\\
  &= C\, e^{s_3 - s_2 - s_1} \int\limits_{\mathbf{q} \in S(3)} e^{\left(\mathbf{q}^T \Lambda_2 \mathbf{q}\right)} d\mathbf{q}\\
  &= C\,e^{s_3-s_2-s_1}\, {}_1F_1\left(2, \tfrac{1}{2}, \Lambda_2\right)
\end{align}
Step 1 of the above is justified by the result from section \ref{scale_change_quat_to_rot}.
The last step comes from recognizing the integral as the normalizing constant of a Bingham distribution on $S(3)$. This normalizing constant is from \cite{jupp1979maximum}.

\section{Pascal3D preprocessing}
\label{appendix_preprocessing}

\begin{figure}[htpb]
    \centering
    \includegraphics[scale=0.2]{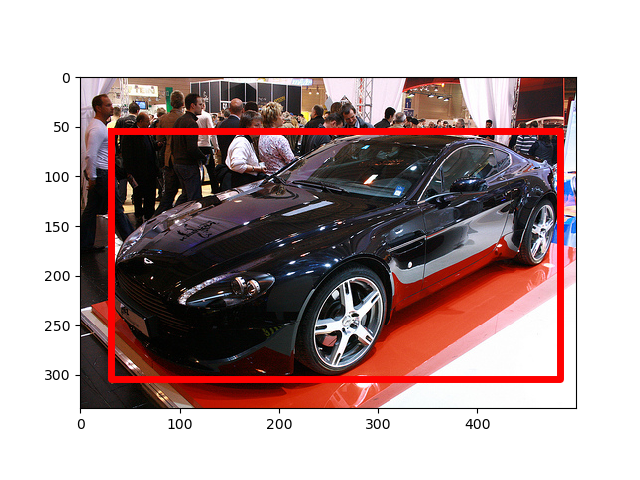}
    \includegraphics[scale=0.3]{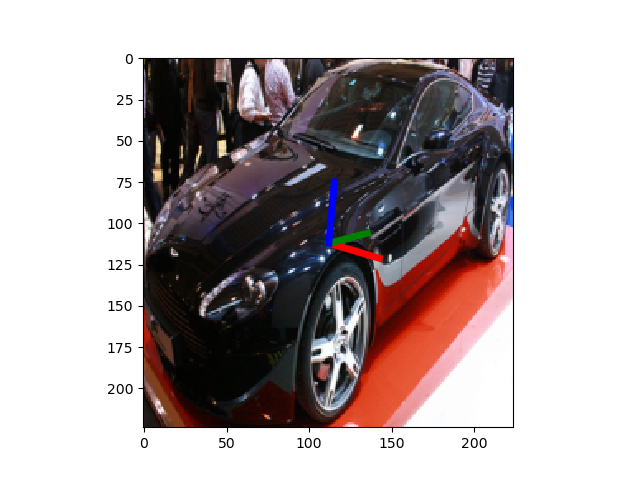}
    \includegraphics[scale=0.3]{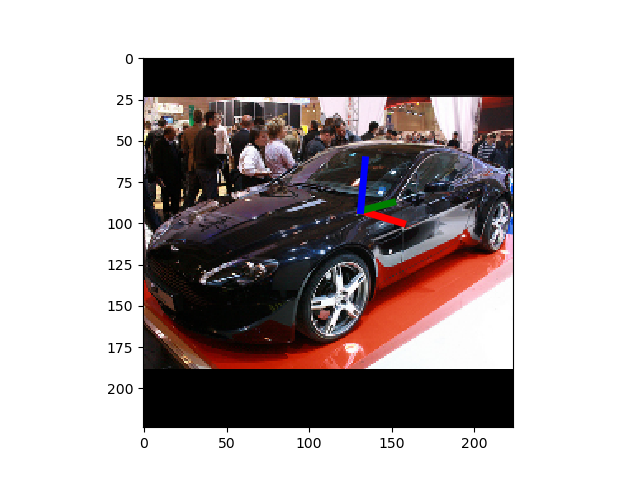}
    \caption{illustration of how naive cropping changes perceived orientation. Left: original image with bounding box. Middle: Cropped image with unchanged ground truth orientation. Right: warped image with adjusted ground truth orientation. As we can see the middle image appears to have a different azimuth due to aspect ratio of bounding box. On the right image the green axis is well aligned with the "backwards" direction of the car.}
    \label{fig:cropping}
\end{figure}

The flaws of using cropping as a method are twofold. Firstly if the width and height of the bounding box are not the same the scaling can cause artifacts which are very similar to a rotation. For reference see figure \ref{fig:cropping}. Secondly as the object moves away from the principal axis of the camera the cropped image will change in a similar manner compared to when the object is rotated.

To solve these issues we assume that the position of the bounding box of the object is known. We now create a desired pinhole camera which is rotated relative to the real camera in such a way that the principal axis is facing the center of the bounding box. We let the intrinsic of the desired camera be
\begin{gather*}
    I_{ideal} = \begin{bmatrix}
    f & 0 & s/2 \\
    0 & f & s/2 \\
    0 & 0 & 1
\end{bmatrix}
\end{gather*}
Where s is the size of the pictures taken with this camera and f is picked to be the largest value such that all points of the bounding box is still inside of the pictures taken with the virtual camera.

The transformation between the desired camera and the real camera is now a homography, we can simulate taking pictures with the desired camera warping the image from the real camera.

When we estimate orientations we estimate them relative to the new camera. This is not cheating since if one wanted the orientation in the camera coordinate system on could just apply the (known) inverse rotation between the two cameras, and the loss and evaluation metric are both invariant to what coordinate system is used.

\subsection{Camera details}
To compute rotation for the desired camera we first backproject every corner of the bounding box onto a sphere by
\begin{gather}
    \hat{p}
    = I_{real}^{-1} \begin{bmatrix}
    p_y\\ p_x\\ 1\\
    \end{bmatrix}
\end{gather}

followed by

\begin{gather}
    p = \dfrac {\hat{p}} {||\hat{p}||_2}
\end{gather}

where $p_x$, $p_y$ is the position in the image and $p$ is the backprojected point on a sphere.

We now have 4 points on a sphere, to find the desired direction of the principal point we apply a modified version of Welzl's algorithm to find the minimal enclosing sphere of these points, subject to the constraint that the center of the enclosing sphere have a center at distance one from the origin.

If the bounding box spans more than 180 degrees one could get a solution from Welzl's algorithm which is pointing 180 degrees in the wrong direction. Since the datasets use normal cameras this will not happen. In addition to this, if the bounding box spans more than 180 degrees it is not possible for a pinhole camera to capture the whole object, due to limitations of the pinhole model.

There is one more degree of freedom, for the rotation of the ideal camera, we eliminate this degree by adding a constraint on the direction for the y axis in the desired image.

We can now choose the focal length to be the largest value constrained by the fact that all bounding box coordinates has to be projected in $[0, s]\times [0, s]$, i.e. visible in the warped image.

%% file: main.bbl
\begin{thebibliography}{34}
\providecommand{\natexlab}[1]{#1}
\providecommand{\url}[1]{\texttt{#1}}
\expandafter\ifx\csname urlstyle\endcsname\relax
  \providecommand{\doi}[1]{doi: #1}\else
  \providecommand{\doi}{doi: \begingroup \urlstyle{rm}\Url}\fi

\bibitem[Ariz et~al.(2016)Ariz, Bengoechea, Villanueva, and
  Cabeza]{ariz2016novel}
Mikel Ariz, Jos{\'e}~J. Bengoechea, Arantxa Villanueva, and Rafael Cabeza.
\newblock A novel 2d/3d database with automatic face annotation for head
  tracking and pose estimation.
\newblock \emph{Computer Vision and Image Understanding}, 148:\penalty0
  201--210, 2016.

\bibitem[Bingham(1974)]{bingham1974antipodally}
Christopher Bingham.
\newblock An antipodally symmetric distribution on the sphere.
\newblock \emph{The Annals of Statistics}, pages 1201--1225, 1974.

\bibitem[Doumanoglou et~al.(2016)Doumanoglou, Balntas, Kouskouridas, and
  Kim]{DBLP:journals/corr/DoumanoglouBKK16}
Andreas Doumanoglou, Vassileios Balntas, Rigas Kouskouridas, and Tae{-}Kyun
  Kim.
\newblock Siamese regression networks with efficient mid-level feature
  extraction for 3d object pose estimation.
\newblock \emph{arXiv preprint arXiv:1607.02257}, 2016.

\bibitem[Downs(1972)]{downs1972orientation}
Thomas~D Downs.
\newblock Orientation statistics.
\newblock \emph{Biometrika}, 59\penalty0 (3):\penalty0 665--676, 1972.

\bibitem[Eggert et~al.(1997)Eggert, Lorusso, and Fisher]{eggert1997estimating}
David~W Eggert, Adele Lorusso, and Robert~B Fisher.
\newblock Estimating 3-d rigid body transformations: a comparison of four major
  algorithms.
\newblock \emph{Machine vision and applications}, 9\penalty0 (5-6):\penalty0
  272--290, 1997.

\bibitem[Gilitschenski et~al.(2019)Gilitschenski, Sahoo, Schwarting, Amini,
  Karaman, and Rus]{gilitschenski2019deep}
Igor Gilitschenski, Roshni Sahoo, Wilko Schwarting, Alexander Amini, Sertac
  Karaman, and Daniela Rus.
\newblock Deep orientation uncertainty learning based on a bingham loss.
\newblock In \emph{International Conference on Learning Representations}, 2019.

\bibitem[Grabner et~al.(2018)Grabner, Roth, and Lepetit]{grabner20183d}
Alexander Grabner, Peter~M Roth, and Vincent Lepetit.
\newblock 3d pose estimation and 3d model retrieval for objects in the wild.
\newblock In \emph{Proceedings of the IEEE Conference on Computer Vision and
  Pattern Recognition}, pages 3022--3031, 2018.

\bibitem[Guo et~al.(2017)Guo, Pleiss, Sun, and Weinberger]{guo2017calibration}
Chuan Guo, Geoff Pleiss, Yu~Sun, and Kilian~Q Weinberger.
\newblock On calibration of modern neural networks.
\newblock In \emph{Proceedings of the 34th International Conference on Machine
  Learning-Volume 70}, pages 1321--1330. JMLR. org, 2017.

\bibitem[Herz(1955)]{herz1955bessel}
Carl~S Herz.
\newblock Bessel functions of matrix argument.
\newblock \emph{Annals of Mathematics}, pages 474--523, 1955.

\bibitem[Jupp et~al.(1979)Jupp, Mardia, et~al.]{jupp1979maximum}
Peter~E Jupp, Kanti~V Mardia, et~al.
\newblock Maximum likelihood estimators for the matrix von mises-fisher and
  bingham distributions.
\newblock \emph{The Annals of Statistics}, 7\penalty0 (3):\penalty0 599--606,
  1979.

\bibitem[Khatri and Mardia(1977)]{Khatri:rss:77}
C.~Khatri and K.~Mardia.
\newblock The von mises-fisher matrix distribution in orientation statistics.
\newblock \emph{Journal of the Royal Statistical Society: Series B},
  39\penalty0 (1):\penalty0 95--106, 1977.

\bibitem[Koev and Edelman(2006)]{koev2006efficient}
Plamen Koev and Alan Edelman.
\newblock The efficient evaluation of the hypergeometric function of a matrix
  argument.
\newblock \emph{Mathematics of Computation}, 75\penalty0 (254):\penalty0
  833--846, 2006.

\bibitem[Lee(2018)]{lee2018bayesian}
Taeyoung Lee.
\newblock Bayesian attitude estimation with the matrix fisher distribution on
  so (3).
\newblock \emph{IEEE Transactions on Automatic Control}, 63\penalty0
  (10):\penalty0 3377--3392, 2018.

\bibitem[Lepetit et~al.(2009)Lepetit, Moreno-Noguer, and Fua]{Lepetit09}
Vincent Lepetit, Franscesc Moreno-Noguer, and Pascal Fua.
\newblock {EPnP: An Accurate O(n) Solution to the P$n$P Problem}.
\newblock \emph{{International Journal of Computer Vision}}, 2009.

\bibitem[Liao(2020)]{spherical_regression_code}
Shuai Liao.
\newblock spherical regression, 2020.
\newblock URL \url{https://github.com/leoshine/Spherical\_Regression}.

\bibitem[Liao et~al.(2019)Liao, Gavves, and Snoek]{liao2019spherical}
Shuai Liao, Efstratios Gavves, and Cees~GM Snoek.
\newblock Spherical regression: Learning viewpoints, surface normals and 3d
  rotations on n-spheres.
\newblock In \emph{Proceedings of the IEEE Conference on Computer Vision and
  Pattern Recognition}, pages 9759--9767, 2019.

\bibitem[Lind{\'e}n et~al.(2018)Lind{\'e}n, Sj{\"o}strand, and
  Proutiere]{linden2018appearance}
Erik Lind{\'e}n, Jonas Sj{\"o}strand, and Alexandre Proutiere.
\newblock Appearance-based 3d gaze estimation with personal calibration.
\newblock \emph{arXiv preprint arXiv:1807.00664}, 1\penalty0 (3):\penalty0 6,
  2018.

\bibitem[Mahendran et~al.(2017)Mahendran, Ali, and Vidal]{mahendran20173d}
Siddharth Mahendran, Haider Ali, and Ren{\'e} Vidal.
\newblock 3d pose regression using convolutional neural networks.
\newblock In \emph{Proceedings of the IEEE International Conference on Computer
  Vision Workshops}, pages 2174--2182, 2017.

\bibitem[Mahendran et~al.(2018)Mahendran, Ali, and Vidal]{mahendran2018mixed}
Siddharth Mahendran, Haider Ali, and Rene Vidal.
\newblock A mixed classification-regression framework for 3d pose estimation
  from 2d images.
\newblock \emph{arXiv preprint arXiv:1805.03225}, 2018.

\bibitem[Marchand et~al.(2016)Marchand, Uchiyama, and
  Spindler]{marchand:hal-01246370}
Eric Marchand, Hideaki Uchiyama, and Fabien Spindler.
\newblock {Pose Estimation for Augmented Reality: A Hands-On Survey}.
\newblock \emph{{IEEE Transactions on Visualization and Computer Graphics}},
  22\penalty0 (12):\penalty0 2633 -- 2651, December 2016.

\bibitem[Mardia and Jupp(2009)]{directional_statistics}
Kanti~V Mardia and Peter~E Jupp.
\newblock \emph{Directional Statistics}.
\newblock John Wiley \& Sons, 2009.

\bibitem[Meng et~al.(2017)Meng, Wang, and Liu]{Meng_2017}
Xiaoli Meng, Heng Wang, and Bingbing Liu.
\newblock A robust vehicle localization approach based on gnss/imu/dmi/lidar
  sensor fusion for autonomous vehicles.
\newblock \emph{Sensors}, 17:\penalty0 2140, Sep 2017.

\bibitem[Pavlakos et~al.(2017)Pavlakos, Zhou, Chan, Derpanis, and
  Daniilidis]{pavlakos20176}
Georgios Pavlakos, Xiaowei Zhou, Aaron Chan, Konstantinos~G Derpanis, and
  Kostas Daniilidis.
\newblock 6-dof object pose from semantic keypoints.
\newblock In \emph{2017 IEEE International Conference on Robotics and
  Automation (ICRA)}, pages 2011--2018. IEEE, 2017.

\bibitem[Prokudin et~al.(2018)Prokudin, Gehler, and Nowozin]{prokudin2018deep}
Sergey Prokudin, Peter Gehler, and Sebastian Nowozin.
\newblock Deep directional statistics: Pose estimation with uncertainty
  quantification.
\newblock In \emph{Proceedings of the European Conference on Computer Vision
  (ECCV)}, pages 534--551, 2018.

\bibitem[Se et~al.(2001)Se, Lowe, and Little]{se2001vision}
Stephen Se, David Lowe, and Jim Little.
\newblock Vision-based mobile robot localization and mapping using
  scale-invariant features.
\newblock In \emph{Proceedings 2001 ICRA. IEEE International Conference on
  Robotics and Automation (Cat. No. 01CH37164)}, volume~2, pages 2051--2058.
  IEEE, 2001.

\bibitem[Su et~al.(2015)Su, Qi, Li, and Guibas]{su2015render}
Hao Su, Charles~R Qi, Yangyan Li, and Leonidas~J Guibas.
\newblock Render for cnn: Viewpoint estimation in images using cnns trained
  with rendered 3d model views.
\newblock In \emph{Proceedings of the IEEE International Conference on Computer
  Vision}, pages 2686--2694, 2015.

\bibitem[Tremblay et~al.(2018)Tremblay, To, Sundaralingam, Xiang, Fox, and
  Birchfield]{tremblay2018deep}
Jonathan Tremblay, Thang To, Balakumar Sundaralingam, Yu~Xiang, Dieter Fox, and
  Stan Birchfield.
\newblock Deep object pose estimation for semantic robotic grasping of
  household objects.
\newblock \emph{arXiv preprint arXiv:1809.10790}, 2018.

\bibitem[Tulsiani and Malik(2015)]{tulsiani2015viewpoints}
Shubham Tulsiani and Jitendra Malik.
\newblock Viewpoints and keypoints.
\newblock In \emph{Proceedings of the IEEE Conference on Computer Vision and
  Pattern Recognition}, pages 1510--1519, 2015.

\bibitem[Weidenbacher et~al.(2006)Weidenbacher, Layher, Bayerl, and
  Neumann]{Weidenbacher2006}
Ulrich Weidenbacher, Georg Layher, Pierre Bayerl, and Heiko Neumann.
\newblock Detection of head pose and gaze direction for human-computer
  interaction.
\newblock In \emph{Perception and Interactive Technologies}, pages 9--19.
  Springer Berlin Heidelberg, 2006.

\bibitem[Wu et~al.(2015)Wu, Song, Khosla, Yu, Zhang, Tang, and Xiao]{wu20153d}
Zhirong Wu, Shuran Song, Aditya Khosla, Fisher Yu, Linguang Zhang, Xiaoou Tang,
  and Jianxiong Xiao.
\newblock 3d shapenets: A deep representation for volumetric shapes.
\newblock In \emph{Proceedings of the IEEE conference on computer vision and
  pattern recognition}, pages 1912--1920, 2015.

\bibitem[Xiang et~al.(2014)Xiang, Mottaghi, and Savarese]{xiang_wacv14}
Yu~Xiang, Roozbeh Mottaghi, and Silvio Savarese.
\newblock Beyond pascal: A benchmark for 3d object detection in the wild.
\newblock In \emph{IEEE Winter Conference on Applications of Computer Vision
  (WACV)}, 2014.

\bibitem[Xiang et~al.(2018)Xiang, Schmidt, Narayanan, and
  Fox]{xiang2017posecnn}
Yu~Xiang, Tanner Schmidt, Venkatraman Narayanan, and Dieter Fox.
\newblock Posecnn: A convolutional neural network for 6d object pose estimation
  in cluttered scenes.
\newblock \emph{Robotics: Science and Systems (RSS)}, 2018.

\bibitem[Zhang et~al.(2018)Zhang, Sugano, and Bulling]{zhang2018revisiting}
Xucong Zhang, Yusuke Sugano, and Andreas Bulling.
\newblock Revisiting data normalization for appearance-based gaze estimation.
\newblock In \emph{Proceedings of the 2018 ACM Symposium on Eye Tracking
  Research \& Applications}, pages 1--9, 2018.

\bibitem[Zhou et~al.(2019)Zhou, Barnes, Lu, Yang, and Li]{zhou2019continuity}
Yi~Zhou, Connelly Barnes, Jingwan Lu, Jimei Yang, and Hao Li.
\newblock On the continuity of rotation representations in neural networks.
\newblock In \emph{Proceedings of the IEEE Conference on Computer Vision and
  Pattern Recognition}, pages 5745--5753, 2019.

\end{thebibliography}
